\documentclass[lettersize,journal]{IEEEtran}

\usepackage{amsmath}
\usepackage{bm}
\usepackage{amssymb}
\usepackage{mathtools}
\usepackage{cite}
\usepackage{graphicx}
\usepackage{epstopdf}
\usepackage{amsthm}
\usepackage{braket}
\usepackage{algorithm}
\usepackage{algorithmic}
\usepackage{siunitx}
\usepackage{url}
\usepackage{booktabs}     
\usepackage{subfigure}  
\usepackage{float}
\usepackage{multirow}
\usepackage{color}
\usepackage[]{changes} 

\usepackage{hyperref}
\usepackage{marginnote}
\usepackage{lipsum}
\usepackage{makeidx}  

\newtheorem{theorem}{Theorem}

\newtheorem{proposition}[theorem]{Proposition}

\definecolor{hmzcolor}{rgb}{0.523, 0.235, 0.625}

\definecolor{red}{rgb}{0.1,0.1,0.8}
\definecolor{blue}{rgb}{0,0,0.8}
\definecolor{green}{rgb}{0,0.4,0}

\newcommand{\change}[2]{}
\newcommand{\lchange}[2]{}

\newcommand{\vx}{\boldsymbol{x}}
\newcommand{\vd}{\boldsymbol{d}}
\newcommand{\vA}{\boldsymbol{A}}
\newcommand{\vB}{\boldsymbol{B}}
\newcommand{\vG}{\boldsymbol{G}}
\newcommand{\vI}{\boldsymbol{I}}

\newcommand{\vS}{\boldsymbol{S}}

\newcommand{\vP}{\boldsymbol{P}}

\newcommand{\vX}{\boldsymbol{X}}
\newcommand{\vY}{\boldsymbol{Y}}
\newcommand{\vZ}{\boldsymbol{Z}}
\newcommand{\valpha}{\boldsymbol{\alpha}}
\newcommand{\vlambda}{\boldsymbol{\lambda}}
\newcommand{\vtheta}{\boldsymbol{\theta}}
\newcommand{\vomega}{\boldsymbol{\omega}}
\newcommand{\vTheta}{\boldsymbol{\Theta}}
\newcommand{\vOmega}{\boldsymbol{\Omega}}
\newcommand{\vzero}{\boldsymbol{0}}

\begin{document}

\clearpage

\twocolumn
\pagenumbering{arabic}
\setcounter{page}{1}
\renewcommand\thefigure{\arabic{figure}} 
\renewcommand\thetable{\arabic{table}} 
\setcounter{figure}{0}
\setcounter{table}{0}

\title{Decentralized Kernel Ridge Regression Based on Data-Dependent Random Feature}

\author{Ruikai~Yang, 
        Fan~He,
        Mingzhen~He,
        Jie~Yang,~\IEEEmembership{Senior Member,~IEEE},
        and Xiaolin~Huang$^\ast$,~\IEEEmembership{Senior Member,~IEEE}
\thanks{This work is jointly supported by Shanghai Municipal Science and Technology Research Program (22511105600), National Natural Science Foundation of China (62376155, 62376153), and Shanghai Municipal Science and Technology Major Project (2021SHZDZX0102).}
\thanks{R. Yang, M. He, J. Yang, and X. Huang are with the Institute of Image Processing and Pattern Recognition, and the MOE Key Laboratory of System Control and Information Processing, Shanghai Jiao Tong University, Shanghai 200240, China. E-mail: \{ruikai.yang, mingzhen\_he, jieyang, xiaolinhuang\}@sjtu.edu.cn.}
\thanks{F. He is with the Department of Electrical Engineering (ESAT), and the STADIUS Center for Dynamical Systems, Signal Processing and Data Analytics, KU Leuven, B-3001 Leuven, Belgium. E-mail: fan.he@esat.kuleuven.be.}
\thanks{$^\ast$Corresponding author.}
}

\markboth{IEEE Transactions on Neural Networks and Learning Systems}%
{Yang \MakeLowercase{\textit{et al.}}: Decentralized Kernel Ridge Regression Based on Data-dependent Random Feature}


\maketitle

\begin{abstract}
Random feature (RF) has been widely used for node consistency in decentralized kernel ridge regression (KRR). Currently, the consistency is guaranteed by imposing constraints on coefficients of features, necessitating that the random features on different nodes are identical. However, in many applications, data on different nodes varies significantly on the number or distribution, which calls for adaptive and data-dependent methods that generate different RFs. To tackle the essential difficulty, we propose a new decentralized KRR algorithm that pursues consensus on decision functions, which allows great flexibility and well adapts data on nodes. The convergence is rigorously given and the effectiveness is numerically verified: by capturing the characteristics of the data on each node, while maintaining the same communication costs as other methods, we achieved an average regression accuracy improvement of 25.5\% across six real-world data sets.
\end{abstract}


\section{Introduction}
\IEEEPARstart{K}{ernel} methods are widely applied and insightfully investigated for their ability to handle complex nonlinear problems into simple linear problems \cite{xu2013soft,mohsenzadeh2015gaussian,hong2022active,he2024diffusion}. For a compact data set $\mathcal{X}\subset \mathbb{R}^d$ in kernel methods, there is usually a nonlinear mapping $\phi:\mathcal{X}\rightarrow \mathbb{R}^P$ ($P$ can be infinite) to transform data from the sample space to a high-dimensional space. Suppose that $\phi$ belongs to a reproducing kernel Hilbert space (RKHS) $\mathcal{H}_k$ that is endowed with a positive definite kernel function $k(\cdot,\cdot):\mathbb{R}^d\times\mathbb{R}^d\rightarrow\mathbb{R}$, where $d$ is the dimension of data. 
Thanks to the kernel trick, one can compute the inner product in the RKHS with a given kernel function without knowing the explicit expression of $\phi$. 
Kernel ridge regression (KRR, \cite{scholkopf2002learning}) is an important non-parametric kernel-based learning method. Given a data set $\mathcal{D}=\lbrace (\vx_i, y_i)\rbrace_{i=1}^N$ and $\vx_i\in\mathcal{X}\subseteq\mathbb{R}^d$, $y_i\in\mathcal{Y}\subseteq\mathbb{R}$, KRR considers the following optimization problem,
\begin{equation}
    \begin{aligned}
        \min_{f\in\mathcal{H}_k} &\quad\frac{1}{N}\sum_{i=1}^N \ell \left( f(\vx_i), y_i\right)+\lambda\left\| f\right\|_{\mathcal{H}_k}^2,
    \end{aligned}
\end{equation}
where $\lambda > 0$ is the regularization parameter and $\ell: \mathbb{R}\times\mathbb{R}\rightarrow\mathbb{R}$ is a given loss function. 

In today's large-scale settings, distributed computing frameworks \cite{zhang2012imapreduce,7935426,8051074,jiang2021local,wang2023consensus} have attracted much attention in many areas. 
Massive distributed sensors and devices, for example, collect data in wireless communication networks \cite{polo2015design} and the Internet of Things \cite{9288933}.
To protect patients' privacy, some biological big data must be dispersed across institutions and intentionally isolated \cite{weber2014finding}. 
The advantages and necessity of distributed computing frameworks have prompted research into distributed KRR (DKRR). Let $J$ be the number of machines or nodes and  $\mathcal{D}=\cup_{j=1}^{J}\mathcal{D}_j$ be divided into $J$ mutually disjoint subsets, where each subset $\mathcal{D}_j=\lbrace (\vx_{j,i},y_{j,i})\rbrace_{i=1}^{N_j}$ $(1\leqslant j\leqslant J)$. DKRR considers the following optimization problem,
\begin{equation}
    \begin{aligned}
        \min_{\lbrace f_j\in\mathcal{H}_k\rbrace_{j=1}^J} &\ \sum_{j=1}^J \left\{ \frac{1}{N} \sum_{i=1}^{N_j}\ell \left( f_j(\vx_{j,i}), y_{j,i}\right)+\lambda_j\| f_j\|_{\mathcal{H}_k}^2\right\} \\
        \mathrm{s.t.} \quad &\ \  f_1 = f_2 =\;\ldots\;= f_J.
    \end{aligned}
    \label{equ: dkrr}
\end{equation}

Distributed algorithms can be categorized as \emph{centralized} and \emph{decentralized} ones.
Compared to centralized KRR algorithms \cite{zhang2015divide,chang2017distributed,lin2020distributed,liu2021effective}, decentralized algorithms do not require a fusion center and thus are more suitable for applications that care about privacy. But since the local function learned from KRR is supported by local data, namely $\{k(\cdot,\vx_{j,i})\}_{i=1}^{N_j}$, imposing constraints on combination coefficients is almost impossible and in fact unreasonable.
To address the aforementioned difficulty, researchers introduced random features (RFs, \cite{rahimi2007random}) into decentralized frameworks recently \cite{bouboulis2017online,miao2019distributed,richards2020decentralised,xu2021coke,khanduri2021decentralized}.
For a kernel function $k(\vx_i,\vx_j)$, RF constructs an explicit mapping $z:\mathbb{R}^d\rightarrow\mathbb{R}^D$ to approximate $k(\vx_i,\vx_j)\approx z(\vx_i)^\top z(\vx_j)$, where the explicit mapping is obtained by sampling from a certain data-independent distribution. 
Furthermore, since the RF mapping is not bijective, it can ensure the security of $\vx_i$ even when nodes communicate $z(\vx_i)$. Compared to some decentralized methods that require transmitting raw data\cite{forero2010consensus, koppel2018decentralized,he2022decentralized,lian2023decentralized}, this approach will provide crucial assistance in domains where data privacy protection is necessary and prohibits cross-node transmission of raw data\cite{abbe2012privacy, wang2020big, zhao2022intelligent, li2023bfod}.
A typical decentralized KRR with RF, namely decentralized kernel learning via ADMM (DKLA)\cite{xu2021coke}, can be written as below, 
\begin{equation}
    \begin{aligned}
        \min_{\lbrace\vtheta_j\in\mathbb{R}^D\rbrace_{j=1}^J}&\ \sum_{j=1}^J \left\lbrace \frac{1}{N} \sum_{i=1}^{N_j}\ell \left( \vtheta_j^\top z(\vx_{j,i}), y_{j,i}\right)+\lambda_j\| \vtheta_j\|_2^2  \right\rbrace \\
        \mathrm{s.t.} \quad &\ \  \vtheta_j = \vtheta_{p},\; p\in\mathcal{N}_j, \;\forall j,
    \end{aligned}
    \label{equ: dekrr_rf}
\end{equation}
where $\mathcal{N}_j$ denotes one-hop neighbors of the $j$-th node.

There has been significant progress on RF in the past ten years; see, e.g., a recent survey \cite{liu2021random}.
Nowadays, one can generate RF adaptive to data, i.e., for specific data, one can use fewer features to achieve better kernel approximation or performance on subsequent tasks. 
Typically, these data-dependent RF (DDRF) methods sample $D_0$ features from a given probability density, score these features by a predefined metric, and then directly select $D$ ($D<D_0$) features with the highest scores \cite{shahrampour2018data}, \cite{shahrampour2019sampling} or resample those features with higher probability \cite{li2019towards}, \cite{liu2020random}. 
The advantages of these methods come from the fact that the features are data-dependent, from which it follows that the corresponding RFs could be different on different nodes. 
But this does not apply to the current RF-based decentralized algorithms\cite{shen2019random,shen2021distributed,hong2021communication,hong2021distributed}, which share the same premise as the model proposed in \cite{xu2021coke}, relying on consensus on feature coefficients. For these algorithms to work, features on different nodes must be identical.
Moreover, when algorithms deal with imbalanced data, it is preferable to use variable $D_j$ because it can improve communication efficiency without damaging the approximation ability. 
Due to the requirement mentioned before, such a setting also cannot be adopted to current decentralized algorithms.

The purpose of this paper is to develop a decentralized KRR algorithm that allows different nodes to use different RFs. 
We propose to align decision functions on different nodes and iteratively solve the problem with a primal-only method, the convergence of which is proven to be controlled by the local penalty parameter.
Most of the computations and communications can be done before the iteration, and only the $\{\vtheta_j\}$ needs to be communicated in iterations. 
Because decision functions rather than coefficients are used to reach consensus, RFs could adapt to the data on nodes.
This can greatly improve the performance, especially when the data size, data distribution, noise intensity, etc., are different among different nodes. 
For example, a heterogeneous setting of responses $\lbrace y_{j,i}\rbrace_{i=1}^{N_j}$ on each node is tested. 
For the \textit{houses} data set, we run tests on a network of $10$ nodes and each node has $4$ neighbors.
When DKLA \cite{xu2021coke} and our algorithm follow the same convergence rate, DKLA uses $D_j=100$ and our algorithm uses only $D_j=20$ to attain a similar error, which significantly reduces the communication costs.
All of the scenarios above are possible, and the suggested algorithm is promising in actual practice.

The main contributions of this paper are:
\begin{itemize}
    \item We propose a decentralized algorithm named DeKRR-DDRF, which allows the features on each node to be inconsistent. It thus provides the possibility for each node to apply data-dependent RF techniques to benefit in both accuracy and efficiency. 
    \item In DeKRR-DDRF, we give a fast and communication-efficient solving method, where the analytical solution can be directly obtained at every step in iterations. 
    We also prove the convergence of our solving method. 
    \item Several numerical experiments with non-IID and imbalanced data are conducted to test DeKRR-DDRF. The results show that our algorithm outperforms others because it can select features based on the node data characteristics and alter the number of node features.
\end{itemize}

\noindent\textbf{Organization and Notations.} This paper is organized as follows. 
In Section~\ref{Section: algo}, after reformulating the decentralized KRR problem and introducing RF methods, we propose our decentralized algorithm with DDRF. 
In Section~\ref{Section: convegence}, we present the convergence conditions of our solving method. 
In Section~\ref{Section: experiments}, we conduct several experiments to demonstrate that our algorithm is more flexible and efficient than other algorithms with the same communication costs. 
In Section~\ref{Section: conclusion}, we give a brief conclusion of our work.

In this paper, $\mathbb{R}$ denotes the set of real numbers. $\mathbb{E}[\cdot]$ denotes the expectation operator. $\|\cdot\|_2$ denotes the Euclidean norm of a vector. $\mathrm{Tr}(\cdot)$, $\vlambda_{max}(\cdot)$ and $\vlambda_{min}(\cdot)$ denote the trace, the largest eigenvalue and the smallest eigenvalue of a square matrix, respectively. Given a matrix $\vA$, $(\vA)_{i,j}$ denotes $(i,j)$-th entry of $\vA$. 
$\vA\succeq\vB$ denotes that $\vA-\vB$ is a positive semi-definite (PSD) matrix.

\section{Decentralized KRR with Data-dependent Random Features}
\label{Section: algo}
\subsection{Problem Formulation}\label{sec:2A}
Consider a connected network with $J$ nodes which can be described as a symmetric undirected graph $\mathcal{G}=\{\mathcal{V},\mathcal{E}\}$, where $\mathcal{V}$ is the set of nodes and its cardinality $|\mathcal{V}|=J$, and $\mathcal{E}\subset \mathcal{V}\times\mathcal{V}$ is the set of edges. 
For the $j$-th node in the network, $\mathcal{N}_j\subset \mathcal{V}$ is the set of its one-hop neighbors and $\hat{\mathcal{N}}_j=\mathcal{N}_j\cup \{j\}$, where $(j,p), (p,j)\in\mathcal{E}, \forall p\in\mathcal{N}_j$. 
In decentralized settings, a node can only communicate with its one-hop neighbors.
In addition, we assume that each node $j$ has a data set $\mathcal{D}_j=\{\vX_j,\vY_j\}$ which is drawn from a data distribution $\rho_j$ on $\mathcal{X}_j\times\mathcal{Y}_j$. Here $\vX_j=[\vx_{j,1}\ \vx_{j,2}\ \ldots\ \vx_{{j,N_j}}]\in\mathbb{R}^{d\times N_j}$, $\vY_j=[y_{j,1}\ y_{j,2}\ \ldots\ y_{{j,N_j}}]\in\mathbb{R}^{1\times N_j}$. 
And we assume a potential total data set $\mathcal{D}=\cup_j\mathcal{D}_j=\{\vX,\vY\}$. Here $\vX=[\vX_{1}\ \vX_{2}\ \ldots\ \vX_{J}]\in\mathbb{R}^{d\times N}$, $\vY=[\vY_{1}\ \vY_{2}\ \ldots\ \vY_{J}]\in\mathbb{R}^{1\times N}$ and we let $N=\sum_{j=1}^J N_j$. 
Throughout, we suppose the decision function $f$ belongs to a RKHS $\mathcal{H}_k$ that is equipped with a positive definite kernel function $k(\cdot,\cdot):\mathbb{R}^d\times\mathbb{R}^d\rightarrow\mathbb{R}$, 
and we use $\braket{\cdot,\cdot}_{\mathcal{H}_k}$ and $\|\cdot\|_{\mathcal{H}_k}$ to denote the inner product and the norm in $\mathcal{H}_k$, respectively. 

In general, decentralized learning solves the following expected risk minimization problem,
\begin{equation}
    \begin{aligned}
        \min_{\lbrace f_j\in\mathcal{H}_k\rbrace_{j=1}^J} &\ \sum_{j=1}^J\mathbb{E}_{\left(\vx_j, y_j\right)\sim\rho_j}\left[\ell\left(f_j(\vx_j),y_j\right)\right] \\
        \mathrm{s.t.} \quad &\ \  f_1 = f_2 =\;\ldots\; = f_J.
    \end{aligned}
\end{equation}
It is impractical to know every distribution $\rho_j$.
Thus, we usually consider the empirical risk with a given data set, i.e.,
\begin{equation}
    \begin{aligned}
        \min_{\lbrace f_j\in\mathcal{H}_k\rbrace_{j=1}^J} &\ \sum_{j=1}^J\frac{N_j}{N}\left\{\frac{1}{N_j}\sum_{i=1}^{N_j}\ell\left(f_j(\vx_{j,i}),y_{j,i}\right)+\lambda_j\|f_j\|_{\mathcal{H}_k}^2\right\} \\
        \mathrm{s.t.} \quad &\ \  f_1 = f_2 =\;\ldots\;= f_J,
    \end{aligned}
    \label{equ: empRisk}
\end{equation}
where the task-dependent loss function $\ell(f(\vx),y)$ measures the difference between the prediction $f(\vx)$ and the response $y$, and the local regularization parameter $\lambda_j$ is used to control the model's complexity and prevent over-fitting. 
Typically, we choose the squared loss $\ell(f(\vx),y)\coloneqq\left(f(\vx)-y\right)^2$ in KRR.

According to the representer theorem \cite{scholkopf2001generalized}, the optimal decision function of (\ref{equ: empRisk}) can be represented as follows,
\begin{equation}
    f_j^\ast(\cdot) = \valpha^\top K(\cdot,\vX),
\end{equation}
where $\valpha\in\mathbb{R}^N$ and $K(\cdot,\vX)=\left[k(\cdot,\vx_{1,1});\;\ldots\;;k(\cdot,\vx_{J,N_J})\right]$. Using the reproducing property $\braket{k(\cdot,\vx),k(\cdot,\vx^{\prime})}_{\mathcal{H}_k}=k(\vx,\vx^{\prime})$, we have $\|f\|_{\mathcal{H}_k}^2=\valpha^\top \boldsymbol{\mathrm{K}}\valpha$ with $(\boldsymbol{\mathrm{K}})_{i,j}=k(\vx_i,\vx_j)$ $(1\leqslant i,j\leqslant N)$. Then with the squared loss, the problem (\ref{equ: empRisk}) can be rewritten as below,
\begin{equation}
    \begin{aligned}
        \min_{\lbrace \valpha_j\in\mathbb{R}^N\rbrace_{j=1}^J} &\ \sum_{j=1}^J\frac{1}{N}\left\|\valpha_j^\top\tilde{\boldsymbol{\mathrm{K}}}_j-\vY\right\|_2^2+\frac{N_j}{N}\lambda_j\valpha_j^\top \boldsymbol{\mathrm{K}} \valpha_j \\
        \mathrm{s.t.} \quad &\ \  \valpha_1 = \valpha_2 =\;\ldots\;= \valpha_J,
    \end{aligned}
    \label{equ: empRiskalpha}
\end{equation}
where $\tilde{\boldsymbol{\mathrm{K}}}_j\coloneqq[K(\vx_{j,1},\vX)\ K(\vx_{j,2},\vX)\ \ldots\ K(\vx_{j,N_j},\vX)]\in\mathbb{R}^{N\times N_j}$. 
In decentralized settings, however, it will face two main problems. 
Firstly, solving (\ref{equ: empRiskalpha}) requires constructing the matrix $\boldsymbol{\mathrm{K}}$ on each node $j$, which calls for the node to receive raw data sets $\{\mathcal{D}_i\}_{i\in\mathcal{V}\backslash\{j\}}$ from all other nodes. 
This is impractical in today's privacy-conscious world. Moreover, because the dimension of $\valpha_j$ is related to the total amount of data, the high computation, storage, and communication costs are unacceptable for large-scale problems.

\subsection{Random Features in Decentralized Settings}
The main idea of RF methods is sampling $\vOmega=\{\vomega_i\}_{i=1}^D$ in a pre-given probability density $p(\vomega)$ to construct an explicit mapping $z: \mathbb{R}^d\rightarrow\mathbb{R}^D$ to approximate a kernel function:
\begin{equation}
    k(\vx,\vx^{\prime})=\braket{\phi(\vx),\phi(\vx^{\prime})}_{\mathcal{H}_k}\approx \braket{z(\vOmega,\vx),z(\vOmega,\vx^{\prime})},
\end{equation}
where $z(\vOmega,\vx)\coloneqq[\psi(\vomega_1,\vx);\psi(\vomega_2,\vx);\ \ldots\ ;\psi(\vomega_D,\vx)]$. The feature mappings $\psi(\vomega,\vx)$ and the probability densities $p(\vomega)$ for some common kernel functions are summarized in \cite{yang2014random}. For example, given the Gaussian kernel function $k(\vx,\vx^{\prime})=\mathrm{exp}\left(\frac{\|\vx-\vx^{\prime}\|_2^2}{2\sigma^2}\right)$, we sample $\vomega$ from $\mathcal{N}(0,\frac{1}{\sigma^2}\vI_d)$, and there are two ways to construct the real-valued mapping $\psi(\vomega_i,\vx)$:
\begin{equation}
    \psi(\vomega_i,\vx)=\frac{1}{\sqrt{D}}[\mathrm{cos}(\vomega_i^\top\vx);\mathrm{sin}(\vomega_i^\top\vx)]
    \label{equ: mapping1}
\end{equation}
or 
\begin{equation}
    \psi(\vomega_i,\vx)=\sqrt{\frac{2}{D}}\mathrm{cos}(\vomega_i^\top\vx+b_i),
    \label{equ: mapping2}
\end{equation}
where  $b_i$ is sampled uniformly from $[0,2\pi]$. These ways of building the mapping are also called random Fourier features (RFFs, \cite{rahimi2007random}). 

The original process of RFF is data-independent and quite straightforward. Data-dependent RF methods could further improve the performance, i.e., lower approximation error with fewer features. Successful attempts include re-weighted RFF \cite{shahrampour2018data}, \cite{sinha2016learning}, \cite{agrawal2019data} and leverage score based RFF \cite{li2019towards}, \cite{liu2020random}.
A typical way, e.g., energy-based RFF \cite{shahrampour2018data}, is first to generate $D_0$ random features and then introduce a score function $S(\vomega)$ to select good features.
Since $S(\vomega)$ is calculated on data, the features selected on each node will differ, resulting in a different explicit mapping.
This renders requiring consensus on the combination coefficients meaningless.

\subsection{Algorithm Development}
With the features selected by DDRF methods, the local decision function $f_j(\cdot)$ can be written as follows, 
\begin{equation}
    f_j(\cdot) =\valpha_j^\top Z_j^\top(\vX) z_j(\cdot) \coloneqq \vtheta_j^\top z_j(\cdot),
\end{equation}
where $Z_j(\vX)\coloneqq[z(\vOmega_j,\vx_1)\ z(\vOmega_j,\vx_2)\ \ldots\ z(\vOmega_j,\vx_N)]\in\mathbb{R}^{D_j\times N}$. 
Then we propose the following decentralized KRR model to get consensus among decision functions on different nodes,  
\begin{equation}
    \begin{aligned}
        \min_{\lbrace\vtheta_j\in\mathbb{R}^{D_j}\rbrace_{j=1}^J}\quad & \sum_{j=1}^{J}\frac{N_j}{N}\left\lbrace\frac{1}{N_j}\left\|\vtheta_j^\top Z_j(\vX_j)-\vY_j\right\|_2^2+\lambda_j\|\vtheta_j\|_2^2\right\rbrace\\
        \mathrm{s.t.}\quad\quad & \vtheta_j^\top z_j(\cdot) = \vtheta_p^\top z_p(\cdot ), \; p\in\hat{\mathcal{N}}_j, \;\forall j.
    \end{aligned}
    \label{the_problem}
\end{equation}
When all nodes utilize the same features, that is, $D_j=D_p$ and $z_j(\cdot)=z_p(\cdot)$, the constraint is equal to pursuing the consistency of $\{\vtheta_j\}$. 
In that case, (\ref{the_problem}) reduces to the model in \cite{xu2021coke}. 
When the feature mappings are data-dependent, they differ between nodes, causing the equality constraint $\vtheta_j^\top z_j(\cdot) = \vtheta_p^\top z_p(\cdot)$ to be excessively strict to satisfy all $\vx\in\mathcal{X}$.
Thus, we consider the following relaxed problem with an objective function $\mathcal{L}\left(\vtheta_1,\vtheta_2,\;\ldots\;,\vtheta_J\right)$, where we use the data on each node and its neighbors to pursue the consistency of decision functions, 
\begin{equation}
   \begin{aligned}
        \min_{\lbrace\vtheta_j\in\mathbb{R}^{D_j}\rbrace_{j=1}^J}\; \sum_{j=1}^{J}\bigg\{\frac{1}{N}\left\|\vtheta_j^\top Z_j(\vX_j)-\vY_j\right\|_2^2+\frac{N_j}{N}\lambda_j\|\vtheta_j\|_2^2 \\
        +\sum_{p\in\hat{\mathcal{N}}_j}\frac{c_{j,p}}{N|\hat{\mathcal{N}}_j|}\left\|\vtheta_j^\top Z_j(\vX_j)-\vtheta_p^\top Z_p(\vX_j)\right\|_2^2\bigg\}.
    \end{aligned}
    \label{equ: objectiveFunction}
\end{equation}
Here the penalty parameters $c_{j,p}$ control the alignment of decision functions on different nodes, and different quadratic terms can be multiplied by different $c_{j,p}$. In the detailed solution and theoretical analysis later, we use two values: 
\begin{equation}
    c_{j,p} \Rightarrow
    \left\{
    \begin{array}{cc}
        c_{j,\mathrm{self}}, & j=p,\\
        c_{j,\mathrm{nei}},&j\neq p.\\
    \end{array}
    \right.
    \label{equ: splitC_p}
\end{equation}

Roughly, the objective function $\mathcal{L}$ is convex w.r.t. $\vTheta\coloneqq[\vtheta_1;\vtheta_2;\;\ldots\; ;\vtheta_J]$ and easy to solve with a fusion center, e.g., using the analytical formulation. 
In decentralized frameworks, however, this would be unfeasible. 
Furthermore, $\mathcal{L}$ cannot be decoupled to each node, which complicates algorithm design in decentralized frameworks.
By ``decouple'', we mean that the objective function can be represented as $\mathcal{L}=\sum_{j=1}^J\mathcal{L}_j$, where $\mathcal{L}_j$ only depends on the information of the $j$-th node.
In the following, we establish a decentralized iterative method to solve (\ref{equ: objectiveFunction}).

Considering that each node can only get information about itself and its one-hop neighbors, we extract the following terms from $\mathcal{L}$ and merge them as 
\begin{equation}
    \begin{aligned}
        &S_j\left(\vtheta_j,\{\vtheta_p\}_{p\in\mathcal{N}_j}\right)\\
        &\coloneqq \  \frac{1}{N}\left\|\vtheta_j^\top Z_j(\vX_j)-\vY_j\right\|_2^2+\frac{N_j\lambda_j}{N}\left\|\vtheta_j\right\|_2^2\\
        &\quad\ \ +\sum_{p\in\hat{\mathcal{N}}_j}\tilde{c}_{j,p}\left\|\vtheta_j^\top Z_j(\vX_j)-\vtheta_p^\top Z_p(\vX_j)\right\|_2^2\\
        &\quad\ \ +\sum_{p\in\hat{\mathcal{N}}_j}\tilde{c}_{p,j}\left\|\vtheta_j^\top Z_j(\vX_p)-\vtheta_p^\top Z_p(\vX_p)\right\|_2^2,
    \end{aligned}
    \label{equ: S_j}
\end{equation}
where $\tilde{c}_{j,p}\coloneqq c_{j,p}/(N|\hat{\mathcal{N}}_j|)$. 
The first two terms in (\ref{equ: S_j}) are only related to the local information, and the last two are coupled with neighbors. 
In iterations, each node locally optimizes the following sub-problem, where $\{\vtheta_p\}_{p\in\mathcal{N}_j}$ are known and are viewed to be constants,
\begin{equation}
    \min_{\vtheta_j}\quad S_j\left(\vtheta_j;\{\vtheta_p\}_{p\in\mathcal{N}_j}\right).
    \label{equ: minSj}
\end{equation}

To well describe the details, we define $\vZ_{i,j}\coloneqq Z_i(\vX_j)\in\mathbb{R}^{D_j\times N_j}$, $\lambda_j\coloneqq\frac{\lambda N}{JN_j}$ and the following local auxiliary matrices,
\begin{equation}
    \begin{aligned}
        \vG_j\coloneqq &\  \Big[\left(1/N+2\tilde{c}_{j,\mathrm{self}}+|\mathcal{N}_j|\tilde{c}_{j,\mathrm{nei}}\right)\vZ_{j,j}\vZ_{j,j}^\top+\frac{\lambda}{J}\vI_{D_j}\\
        &\quad+\sum_{p\in\mathcal{N}_j}\tilde{c}_{p,\mathrm{nei}}\vZ_{j,p}\vZ_{j,p}^\top\Big]^{-1}\in\mathbb{R}^{D_j\times D_j},\\
        \vd_j\coloneqq &\ \frac{1}{N}\vZ_{j,j}\vY_j^\top\in\mathbb{R}^{D_j},\\
        \vS_j\coloneqq &\ 2\tilde{c}_{j,\mathrm{self}}\vZ_{j,j}\vZ_{j,j}^\top\in\mathbb{R}^{D_j\times D_j},\\
        \vP_{j,p}\coloneqq&\ \tilde{c}_{j,\mathrm{nei}}\vZ_{j,j}\vZ_{p,j}^\top+\tilde{c}_{p,\mathrm{nei}}\vZ_{j,p}\vZ_{p,p}^\top\in\mathbb{R}^{D_j\times D_p}.
    \end{aligned}
    \label{equ: auxMat}
\end{equation}
Note that these auxiliary matrices can be constructed by a finite number of communications with one-hop neighbors before the iteration and remain unchanged throughout our algorithm, which is helpful to simplify the calculation process. 
Due to the use of RF methods, the communication only involves $Z_{i}(\vX_{j})$ $((i,j)\in\mathcal{E})$, which ensures data privacy.
Assuming that at time $k$, the $j$-th node has received $\{\vtheta_p^k\}_{p\in\mathcal{N}_j}$. And then we locally update $\vtheta_j$ by solving $\nabla_{\vtheta_j} S_j=0$ in (\ref{equ: minSj}), which yields
\begin{equation}
    \begin{aligned}
        \vtheta_j^{k+1} = & \bigg[\left(\frac{1}{N}+2\tilde{c}_{j,\mathrm{self}}+|\mathcal{N}_j|\tilde{c}_{j,\mathrm{nei}}\right)\vZ_{j,j}\vZ_{j,j}^\top+\frac{\lambda}{J} \vI_{D_j}\\
        +\sum_{p\in\mathcal{N}_j}&\tilde{c}_{p,\mathrm{nei}}\vZ_{j,p}\vZ_{j,p}^\top\bigg]^{-1}\bullet\bigg[\frac{1}{N}\vZ_{j,j}\vY_j^\top+2\tilde{c}_{j,\mathrm{self}}\vZ_{j,j}\vZ_{j,j}^\top\vtheta_j^k\\
        +\sum_{p\in\mathcal{N}_j}&\left(\tilde{c}_{j,\mathrm{nei}}\vZ_{j,j}\vZ_{p,j}^\top+\tilde{c}_{p,\mathrm{nei}}\vZ_{j,p}\vZ_{p,p}^\top\right)\vtheta_p^{k}\bigg].
    \end{aligned}
    \label{equ: update}
\end{equation}

\noindent By the local auxiliary matrices in (\ref{equ: auxMat}), equation above can be further reduced to
\begin{equation}
    \begin{aligned}
        \vtheta_j^{k+1} = \vG_j\Big(\vd_j+\vS_j\vtheta_j^k +\sum_{p\in\mathcal{N}_j}\vP_{j,p}\vtheta_p^{k}\Big).
    \end{aligned}
    \label{equ: updateSimplify}
\end{equation}
The above algorithm is summarized in Algorithm~\ref{algo: ours}.

For a distributed method, we care about the computation and communication costs. 
Due to the simplicity of our solving method, the costs can be easily given below,  

\begin{enumerate}
    \item The main computational costs arise from calculating (\ref{equ: auxMat}) and (\ref{equ: updateSimplify}).  For the former, the computational complexities of constructing the four types of auxiliary matrices $\vG_j$, $\vd_j$, $\vS_j$, and $\{\vP_{j,p}\}_{p\in\mathcal{N}_j}$ on each node, are $\mathcal{O}(D_j^3)$, $\mathcal{O}(D_jN_j)$, $\mathcal{O}(D_j^2N_j)$, and $\mathcal{O}(D_jD_p\max\{N_j,N_p\})$, respectively. As for the latter, the required computational complexity is $\mathcal{O}(\max\{D_j^2,D_j\max_{p\in\mathcal{N}_j}\{D_p\}\})$. For the sake of simplification, we assume that the mean of all $\{D_j\}_{j=1}^J$ is $\bar{D}$. Therefore, the overall computational complexity is dominated by
    $$\mathcal{O}\left(\max\left\{\bar{D}^3,\bar{D}^2N_j\right\}\right).$$ 
    \item In iterations, our method only needs to communicate $\{\vtheta_j\}$. For the whole network, the communication costs are proportional to
    $$\sum_{j=1}^J|\mathcal{N}_j|D_j.$$
    When the network topology is fixed, reducing $D_j$ on each node can lower total communication costs, which reflects the benefit of adopting DDRF.
\end{enumerate}

\begin{algorithm}[tb]
    \caption{Decentralized KRR with Data-dependent RF}
    \label{algo: ours}
\begin{algorithmic}[1]
    \REQUIRE the data set $\{\mathcal{D}_j\}_{j=1}^J$, the shift-invariant kernel $k$, the number of random features $\{D_j\}_{j=1}^J$, the regularization parameter $\lambda$, and the penalty parameters $c_{j,p}$.
    \ENSURE the decision function $\{f_j(\cdot)=\vtheta_j^\top Z_j(\cdot)\}_{j=1}^J$.
    \STATE \textbf{Initialize:} $\vtheta_j^{(0)} = \boldsymbol{0}$.
    \STATE \textbf{For all local nodes $j=1,\cdots,J$, do in parallel:}
    \STATE Utilize a data-dependent random feature method to select $D_j$ features.
    \STATE Send $\{\vomega_{j,i}\}_{i=1}^{D_j}$ (using (\ref{equ: mapping1})) or $\{\vomega_{j,i},b_{j,i}\}_{i=1}^{D_j}$ (using (\ref{equ: mapping2})) and $Z_j(\vX_j)$ to one-hop neighbors .
    \STATE Receive $\{\vomega_{p,i}\}_{i=1}^{D_p}$ or $\{\vomega_{p,i},b_{p,i}\}_{i=1}^{D_p}$ and $\{Z_p(\vX_p)\}_{p\in\mathcal{N}_j}$ from one-hop neighbors.
    \STATE Compute $\{Z_p(\vX_j)\}_{p\in\mathcal{N}_j}$ and send them to one-hop neighbors.
    \STATE Construct $\vG_j$, $\vd_j$, $\vS_j$, $\{\vP_{j,p}\}_{p\in\mathcal{N}_j}$ by (\ref{equ: auxMat}).
    \STATE Set $k=0$.
    \REPEAT 
    \STATE Send $\vtheta_{j}^k$ to one-hop neighbors.
    \STATE Receive $\{\vtheta_p^k\}_{p\in\mathcal{N}_j}$ from one-hop neighbors.
    \STATE Update $\vtheta_j^{k+1}$ by (\ref{equ: updateSimplify}).
    \STATE $k=k+1$.
    \UNTIL the stop criteria is satisfied.
\end{algorithmic}
\end{algorithm}

\section{Convergence Analysis}
\label{Section: convegence}
In this section, we give the convergence analysis for Algorithm~\ref{algo: ours}.
The challenge stems from the fact that the update to each $\vtheta_j$ is performed in parallel instead of sequentially as in \cite{tseng2001convergence}. 
For a coupled objective function, updating $\vtheta_j$ by (\ref{equ: minSj}) intuitively only ensures $\mathcal{L}(\vtheta_j^{k+1},\{\vtheta_i^k\}_{i\in\mathcal{V}\backslash \{j\}})\leqslant\mathcal{L}(\vtheta_j^{k},\{\vtheta_i^k\}_{i\in\mathcal{V}\backslash \{j\}})$. 
However, when the $\{\vtheta_i^k\}_{i\in\mathcal{V}\backslash \{j\}}$ on the left side are replaced by their updates $\{\vtheta_i^{k+1}\}_{i\in\mathcal{V}\backslash \{j\}}$, it is not clear whether the objective function value is still decreasing.
Therefore, we present the following proposition.

\begin{proposition}
Supposing that 
\begin{equation}
    \tilde{c}_{j,\mathrm{self}}\geqslant\frac{|\mathcal{N}_j|\tilde{c}_{j,\mathrm{nei}}}{2}+\frac{\vlambda_{\mathrm{max}}\left(\sum_{p\in\mathcal{N}_j}\tilde{c}_{p,\mathrm{nei}}\vZ_{j,p}\vZ_{j,p}^\top\right)}{2\vlambda_{\mathrm{min}}\left(\vZ_{j,j}\vZ_{j,j}^\top\right)}
    \label{equ: cself}
\end{equation}
on each node, the objective value of (\ref{equ: objectiveFunction}) is decreasing, i.e., 
\begin{equation}
    \mathcal{L}\left(\vtheta_1^{k+1},\vtheta_2^{k+1},\;\ldots\;,\vtheta_J^{k+1}\right) - \mathcal{L}\left(\vtheta_1^{k},\vtheta_2^{k},\;\ldots\;,\vtheta_J^{k}\right)\leqslant 0.
    \label{equ: Proposition1}
\end{equation}
\label{proposition: convegence}
\end{proposition}
\vspace{-2em}
Proposition~\ref{proposition: convegence} points out that when penalty parameters $\tilde{c}_{j,\mathrm{self}}$ are chosen large enough, the value of the objective function decreases monotonically. 
Since $\mathcal{L}(\vTheta)$ is a convex function, it follows that $\{\mathcal{L}(\vTheta^k)\}$ possesses a lower bound and converges.
On the other hand, the right-hand side of (\ref{equ: cself}) is a theoretical lower bound. In practice, one can start $\tilde{c}_{j,\mathrm{self}}$ with a small value and gradually increase it as the number of iterations increases.

\vspace{1em}
\noindent\textit{Proof of Proposition~\ref{proposition: convegence}.}
First, recalling the definition of $S_j\left(\vtheta_j;\{\vtheta_p\}_{p\in\mathcal{N}_j}\right)$ in (\ref{equ: S_j}) and $\mathcal{L}\left(\vtheta_1,\vtheta_2,\;\ldots\;,\vtheta_J\right)$ in (\ref{equ: objectiveFunction}), we can get the following two expressions:
\begin{equation}
    \begin{aligned}
        &\sum_{j=1}^{J}S_j\left(\vtheta_j^{k+1};\{\vtheta_p^k\}_{p\in\mathcal{N}_j}\right)\\
        &=\sum_{j=1}^J\left(\frac{1}{N}\left\|\left(\vtheta_j^{k+1}\right)^\top\vZ_{j,j}-\vY_j\right\|_2^2+\frac{\lambda}{J}\left\|\vtheta_j^{k+1}\right\|_2^2\right)\\
        &\quad\ +\sum_{j=1}^J\sum_{p\in\hat{\mathcal{N}}_j}\tilde{c}_{j,p}\left\|\left(\vtheta_j^{k+1}\right)^\top\vZ_{j,j}-\left(\vtheta_p^{k}\right)^\top\vZ_{p,j}\right\|_2^2\\
        &\quad\ +\sum_{j=1}^J\sum_{p\in\hat{\mathcal{N}}_j}\tilde{c}_{p,j}\left\|\left(\vtheta_j^{k+1}\right)^\top\vZ_{j,p}-\left(\vtheta_p^{k}\right)^\top\vZ_{p,p}\right\|_2^2
    \end{aligned}
    \label{sum_S}
\end{equation}
and
\begin{equation}
    \begin{aligned}
        &\mathcal{L}\left(\vtheta_1^{k+1},\vtheta_2^{k+1},\;\ldots\;,\vtheta_J^{k+1}\right)
        \\
        &=\sum_{j=1}^J\left(\frac{1}{N}\left\|\left(\vtheta_j^{k+1}\right)^\top\vZ_{j,j}-\vY_j\right\|_2^2+\frac{\lambda}{J}\left\|\vtheta_j^{k+1}\right\|_2^2\right)\\
        &\quad\ +\sum_{j=1}^J\sum_{p\in\hat{\mathcal{N}}_j}\tilde{c}_{j,p}\left\|\left(\vtheta_j^{k+1}\right)^\top\vZ_{j,j}-\left(\vtheta_p^{k+1}\right)^\top\vZ_{p,j}\right\|_2^2.
    \end{aligned}
    \label{L_Theta}
\end{equation}
We suppose $S_j\left(\vtheta_j;\{\vtheta_p\}_{p\in\mathcal{N}_j}\right)$ is a $\mu_j$-strong convex function w.r.t $\vtheta_j$, and then the following inequality holds:
\begin{equation}
    \begin{aligned}
    &S_j\left(\vtheta_j^{k+1};\{\vtheta_p^k\}_{p\in\mathcal{N}_j}\right)-S_j\left(\vtheta_j^k;\{\vtheta_p^k\}_{p\in\mathcal{N}_j}\right)\\
    &\leqslant-\frac{\mu_j}{2}\left\|\vtheta_j^k-\vtheta_j^{k+1}\right\|_2^2\\
    &\leqslant\ 0.
    \end{aligned}
    \label{strong_convex}
\end{equation}
Summing the expressions for all nodes on both sides of (\ref{strong_convex}) and substituting in (\ref{sum_S}) and (\ref{L_Theta}), we obtain the following inequality:
\begin{equation}
    \begin{aligned}
        &\sum_{j=1}^{J}S_j\left(\vtheta_j^{k+1};\{\vtheta_p^k\}_{p\in\mathcal{N}_j}\right) - \sum_{j=1}^{J}S_j\left(\vtheta_j^k;\{\vtheta_p^k\}_{p\in\mathcal{N}_j}\right)\\
        &=\mathcal{L}\left(\vtheta_1^{k+1},\vtheta_2^{k+1},\;\ldots\;,\vtheta_J^{k+1}\right) - \mathcal{L}\left(\vtheta_1^{k},\vtheta_2^{k},\;\ldots\;,\vtheta_J^{k}\right)\\
        &\quad\Bigg\{- \sum_{j=1}^J\sum_{p\in\hat{\mathcal{N}}_j}\tilde{c}_{j,p}\left\|\left(\vtheta_j^{k+1}\right)^\top\vZ_{j,j}-\left(\vtheta_p^{k+1}\right)^\top\vZ_{p,j}\right\|_2^2\\
        &\quad\ \ +\sum_{j=1}^J\sum_{p\in\hat{\mathcal{N}}_j}\tilde{c}_{j,p}\left\|\left(\vtheta_j^{k+1}\right)^\top\vZ_{j,j}-\left(\vtheta_p^{k}\right)^\top\vZ_{p,j}\right\|_2^2\\
        &\quad\ \ +\sum_{j=1}^J\sum_{p\in\hat{\mathcal{N}}_j}\tilde{c}_{p,j}\left\|\left(\vtheta_j^{k+1}\right)^\top\vZ_{j,p}-\left(\vtheta_p^{k}\right)^\top\vZ_{p,p}\right\|_2^2\\
        &\quad\ \ -\sum_{j=1}^J\sum_{p\in\hat{\mathcal{N}}_j}\tilde{c}_{p,j}\left\|\left(\vtheta_j^{k}\right)^\top\vZ_{j,p}-\left(\vtheta_p^{k}\right)^\top\vZ_{p,p}\right\|_2^2\Bigg\}\\
        &\leqslant \sum_{j=1}^J -\frac{\mu_j}{2}\left\|\vtheta_j^{k}-\vtheta_j^{k+1}\right\|_2^2\\
        &\leqslant 0. 
    \end{aligned}
    \label{equ: spread}
\end{equation}
Referring to the operation in (\ref{equ: splitC_p}), we split $\tilde{c}_{j,p}$ into terms w.r.t. $p=j$ and terms w.r.t. $p\in\mathcal{N}_j$ and define them as $\tilde{c}_{j,\mathrm{self}}$ and $\tilde{c}_{j,\mathrm{nei}}$, respectively. Hence the terms between the two braces in (\ref{equ: spread}) can be rewritten as
\begin{equation}
    \begin{aligned}
        &- \sum_{j=1}^J\tilde{c}_{j,\mathrm{nei}}\sum_{p\in\mathcal{N}_j}\left\|\left(\vtheta_j^{k+1}\right)^\top\vZ_{j,j}-\left(\vtheta_p^{k+1}\right)^\top\vZ_{p,j}\right\|_2^2\\
        &+\sum_{j=1}^J\tilde{c}_{j,\mathrm{nei}}\sum_{p\in\mathcal{N}_j}\left\|\left(\vtheta_j^{k+1}\right)^\top\vZ_{j,j}-\left(\vtheta_p^{k}\right)^\top\vZ_{p,j}\right\|_2^2\\
        &+\sum_{j=1}^J\sum_{p\in\mathcal{N}_j}\tilde{c}_{p,\mathrm{nei}}\left\|\left(\vtheta_j^{k+1}\right)^\top\vZ_{j,p}-\left(\vtheta_p^{k}\right)^\top\vZ_{p,p}\right\|_2^2\\
        &-\sum_{j=1}^J\sum_{p\in\mathcal{N}_j}\tilde{c}_{p,\mathrm{nei}}\left\|\left(\vtheta_j^{k}\right)^\top\vZ_{j,p}-\left(\vtheta_p^{k}\right)^\top\vZ_{p,p}\right\|_2^2\\
        &+\sum_{j=1}^J2\tilde{c}_{j,\mathrm{self}}\left\|\left(\vtheta_j^{k+1}\right)^\top\vZ_{j,j}-\left(\vtheta_j^{k}\right)^\top\vZ_{j,j}\right\|_2^2.
    \end{aligned}
    \label{equ: rewritten}
\end{equation}
We know from the symmetry of undirected graphs that if there is a term w.r.t. $(j,p)$, then there must also be a term w.r.t. $(p,j)$ $(j,p\in[J])$. By expanding and merging terms w.r.t. $(j,p)$ and $(p,j)$, (\ref{equ: rewritten}) is transformed into
\begin{equation}
    \begin{aligned}
        &\sum_{(j,p)\in\mathcal{\tilde{E}}}\left(\Delta\vtheta_p^k\right)^\top\left(2\tilde{c}_{j,\mathrm{nei}}\vZ_{p,j}\vZ_{j,j}^\top+2\tilde{c}_{p,\mathrm{nei}}\vZ_{p,p}\vZ_{j,p}^\top\right)\left(\Delta\vtheta_j^k\right)\\
        &\quad+\sum_{j=1}^J\left(\Delta\vtheta_j^k\right)^\top\left(2\tilde{c}_{j,\mathrm{self}}\vZ_{j,j}\vZ_{j,j}^\top\right)\left(\Delta\vtheta_j^k\right)\\
        &=\sum_{(j,p)\in\mathcal{E}}\left(\Delta\vtheta_p^k\right)^\top\left(\tilde{c}_{j,\mathrm{nei}}\vZ_{p,j}\vZ_{j,j}^\top+\tilde{c}_{p,\mathrm{nei}}\vZ_{p,p}\vZ_{j,p}^\top\right)\left(\Delta\vtheta_j^k\right)\\
        &\quad+\sum_{j=1}^J\left(\Delta\vtheta_j^k\right)^\top\left(2\tilde{c}_{j,\mathrm{self}}\vZ_{j,j}\vZ_{j,j}^\top\right)\left(\Delta\vtheta_j^k\right),
    \end{aligned}
    \label{equ: neiandself}
\end{equation}
where $\tilde{\mathcal{E}}$ is the set of edges when graph $\mathcal{G}$ is assumed to be a directed graph and $\Delta\vtheta_j^k\coloneqq \vtheta_j^{k+1}-\vtheta_j^{k}$ is defined for brevity.
To further simplify the expression, we introduce a block matrix $\mathcal{A}$ and a diagonal block matrix $\mathcal{U}$. The $(m,n)$-th block $((m,n)\in\mathcal{V}\times\mathcal{V})$ of $\mathcal{A}$ is defined as
\begin{equation}
    \left[\mathcal{A}\right]_{m,n} \coloneqq 
    \left\{
    \begin{array}{cc}
        \begin{array}{c}
        \begin{aligned}
             &\tilde{c}_{n,\mathrm{nei}}\vZ_{m,n}\vZ_{n,n}^\top\\
             +&\tilde{c}_{m,\mathrm{nei}}\vZ_{m,m}\vZ_{n,m}^\top
        \end{aligned}
        \end{array},
    &(m,n)\in\mathcal{E},\\
    \boldsymbol{0},&\mathrm{otherwise},\\
    \end{array}
    \right.
    \label{equ: A}
\end{equation}
and each diagonal block of $\mathcal{U}$ is defined as 
\begin{equation}
    \left[\mathcal{U}\right]_{j,j}\coloneqq 2\tilde{c}_{j,\mathrm{self}}\vZ_{j,j}\vZ_{j,j}^\top,\quad j\in\mathcal{V}.
    \label{equ: U}
\end{equation}
With (\ref{equ: A}) and (\ref{equ: U}), we can reduce (\ref{equ: neiandself}) to the following,
\begin{equation}
    \left(\vTheta^{k+1}-\vTheta^{k}\right)^\top \left(\mathcal{A}+\mathcal{U}\right) \left(\vTheta^{k+1}-\vTheta^{k}\right),
    \label{equ: thetaAUtheta}
\end{equation}
where $\vTheta^k\coloneqq[\vtheta_1^k;\vtheta_2^k;\;...\; ;\vtheta_J^k]\in\mathbb{R}^{D_s}$ and $D_s\coloneqq \sum_{j=1}^J D_j$. 
Substituting (\ref{equ: thetaAUtheta}) into (\ref{equ: spread}) yields
\begin{equation}
    \begin{aligned}
        &\mathcal{L}\left(\vtheta_1^{k+1},\vtheta_2^{k+1},\;\ldots\;,\vtheta_J^{k+1}\right) - \mathcal{L}\left(\vtheta_1^{k},\vtheta_2^{k},\;\ldots\;,\vtheta_J^{k}\right)\\
        &\leqslant -\left(\vTheta^{k+1}-\vTheta^{k}\right)^\top \left(\mathcal{A}+\mathcal{U}\right) \left(\vTheta^{k+1}-\vTheta^{k}\right).
        \label{equ: L<=A+U}
    \end{aligned}
\end{equation}
To prove (\ref{equ: Proposition1}), we only need to show that $\mathcal{A}+\mathcal{U}$ is a PSD matrix, i.e., $\mathcal{A}+\mathcal{U}\succeq \vzero$.
We rearrange the terms in $\mathcal{A}+\mathcal{U}$ and split them into two parts:
\begin{equation}
    \mathcal{A}+\mathcal{U}=\left(\sum_{j=1}^J\tilde{c}_{j,\mathrm{nei}}\mathcal{W}_j\mathcal{W}_j^\top\right)+\mathcal{S}.
    \label{equ: A+U=W+S}
\end{equation}
For the first part, each block of $\mathcal{W}_j$ is defined as 
\begin{equation}
    \left[\mathcal{W}_j\right]_{m,n} \coloneqq
    \left\{
    \begin{array}{cc}
    \vZ_{j,j},&m=j\ \mathrm{and}\ n\in\mathcal{N}_j,\\
    \vZ_{m,j},&m=n\ \mathrm{and}\ m\in\mathcal{N}_j,\\
    \boldsymbol{0},&\mathrm{otherwise}.
    \end{array}
    \right.
\end{equation}
Apparently, the first part is PSD because 
\begin{equation}
    \begin{aligned}
        \vx^\top\left(\sum_{j=1}^J\tilde{c}_{j,\mathrm{nei}}\mathcal{W}_j\mathcal{W}_j^\top\right)\vx=\sum_{j=1}^J\tilde{c}_{j,\mathrm{nei}}\vx^\top\mathcal{W}_j\mathcal{W}_j^\top\vx\\
        =\sum_{j=1}^J\tilde{c}_{j,\mathrm{nei}}\|\mathcal{W}_j^\top\vx\|_2^2\geqslant 0,\quad \forall\vx\in\mathbb{R}^{D_s}.
    \end{aligned}
    \label{equ: Wx>0}
\end{equation}
The second part $\mathcal{S}$ is a block diagonal matrix, and each of its blocks is
\begin{equation}
    \begin{aligned}
        \left[\mathcal{S}\right]_{j,j} \coloneqq & \ \left[\mathcal{U}\right]_{j,j}-|\mathcal{N}_j|\tilde{c}_{j,\mathrm{nei}}\vZ_{j,j}\vZ_{j,j}^\top\\
        &-\sum_{p\in\mathcal{N}_j}\tilde{c}_{p,\mathrm{nei}}\vZ_{j,p}\vZ_{j,p}^\top\\
        =&\left(2\tilde{c}_{j,\mathrm{self}}-|\mathcal{N}_j|\tilde{c}_{j,\mathrm{nei}}\right)\vZ_{j,j}\vZ_{j,j}^\top\\
        &-\sum_{p\in\mathcal{N}_j}\tilde{c}_{p,\mathrm{nei}}\vZ_{j,p}\vZ_{j,p}^\top,\quad j\in\mathcal{V}.
    \end{aligned}
\end{equation}
All that remains is to determine when $\mathcal{S}$ is also PSD.
We consider its sufficient condition that every diagonal block $[\mathcal{S}]_{j,j}$ is PSD.
When the following condition is satisfied,
\begin{equation}
    \begin{aligned}
        (2\tilde{c}_{j,\mathrm{self}}-|\mathcal{N}_j|\tilde{c}_{j,\mathrm{nei}})\vlambda_{\mathrm{min}}\left(\vZ_{j,j}\vZ_{j,j}^\top\right)\\
        \geqslant \vlambda_{\mathrm{max}}\left(\sum_{p\in\mathcal{N}_j}\tilde{c}_{p,\mathrm{nei}}\vZ_{j,p}\vZ_{j,p}^\top\right),
    \end{aligned}
\end{equation}
which is equivalent to 
\begin{equation}
    \tilde{c}_{j,\mathrm{self}}\geqslant\frac{|\mathcal{N}_j|\tilde{c}_{j,\mathrm{nei}}}{2}+\frac{\vlambda_{\mathrm{max}}\left(\sum_{p\in\mathcal{N}_j}\tilde{c}_{p,\mathrm{nei}}\vZ_{j,p}\vZ_{j,p}^\top\right)}{2\vlambda_{\mathrm{min}}\left(\vZ_{j,j}\vZ_{j,j}^\top\right)},
\end{equation}
$[\mathcal{S}]_{j,j}\succeq \vzero$ is established and consequently $\mathcal{A}+\mathcal{U}\succeq \vzero$ holds. From (\ref{equ: L<=A+U}) we have
\begin{equation}
    \begin{aligned}
        &\mathcal{L}\left(\vtheta_1^{k+1},\vtheta_2^{k+1},\;\ldots\;,\vtheta_J^{k+1}\right) - \mathcal{L}\left(\vtheta_1^{k},\vtheta_2^{k},\;\ldots\;,\vtheta_J^{k}\right)\\
        &\leqslant -\left(\vTheta^{k+1}-\vTheta^{k}\right)^\top \left(\mathcal{A}+\mathcal{U}\right) \left(\vTheta^{k+1}-\vTheta^{k}\right)\\
        &\leqslant 0.
    \end{aligned}
\end{equation}
This completes the proof that the objective value is decreasing with each iteration. 
\qed

\section{Numerical Experiments}
\label{Section: experiments}
In this section, we will evaluate our algorithm in non-IID and imbalanced data settings, which are the two main scenarios where the features on different nodes should differ. 
All experiments are implemented in MATLAB and run on a machine with Intel$^\circledR$ Core$^\text{TM}$ i7-11700KF CPU (3.60 GHz) and 32GB RAM. The source code is available in \url{https://github.com/Yruikk/DeKRR-DDRF}.
\subsection{Experimental Settings}
\noindent\textbf{Metric.} In our experiments, algorithms are evaluated by the relative square error (RSE) defined below, 
\begin{equation*}
    \mathrm{RSE}=\frac{\sum_{i=1}^N\left(f(\vx_i)-y_i\right)^2}{\sum_{i=1}^N\left(y_i-\Bar{y}\right)^2},
\end{equation*}
where $y_i$ is the observation, $\Bar{y}$ is the mean of $\{y_i\}_{i=1}^N$, and $f(\vx_i)$ is the prediction. 
All experiments are repeated 10 times and the mean RSE is reported. 

\noindent\textbf{Data sets and preprocessing.} 
We choose six real-world data sets from libsvm \cite{chang2011libsvm} and UCI \cite{dheeru2017uci}, including $\textit{houses}$, $\textit{air quality}$, $\textit{energy}$, $\textit{twitter}$, $\textit{Tom's hardware}$, and $\textit{wave}$.
The details of the data sets are shown in Tab.~\ref{Tab: data}.
For preprocessing, we normalize the data such that $\vx_{i}$ are scaled to $[0,1]$ and $y_{i}$ are scaled to $[-1,1]$. 
Each node is trained with half of the local data and tested with the remaining half ($N_{j,\mathrm{train}} = N_{j,\mathrm{test}} = N_j/2$).

\begin{table}[htbp]
\centering
\caption{Data sets statistics: $d$ and $N$ denote the number of data dimensions and total data, respectively.}
\begin{tabular}{c|c|c}
\toprule[1.5pt] 
Data sets & $d$ & $N$  \\ \midrule[1pt]
Houses                     & 8                  & 20640            \\
Air Quality                & 13                 & 9357             \\
Energy                     & 27                 & 19735            \\
Twitter                    & 77                 & 98704            \\ 
Tom's Hardware             & 96                 & 29179            \\
Wave                       & 148                & 63600            \\
\bottomrule[1.5pt]
\end{tabular}
\label{Tab: data}
\end{table}

\begin{figure*}[htbp]
	\centering  
	\vspace{0cm} 
	\subfigtopskip=2pt 
	\subfigbottomskip=2pt 
	\subfigcapskip=-2pt 
	\subfigure[Houses]{
		\label{Fig: nonIID.y.1}
		\includegraphics[width=0.31\linewidth]{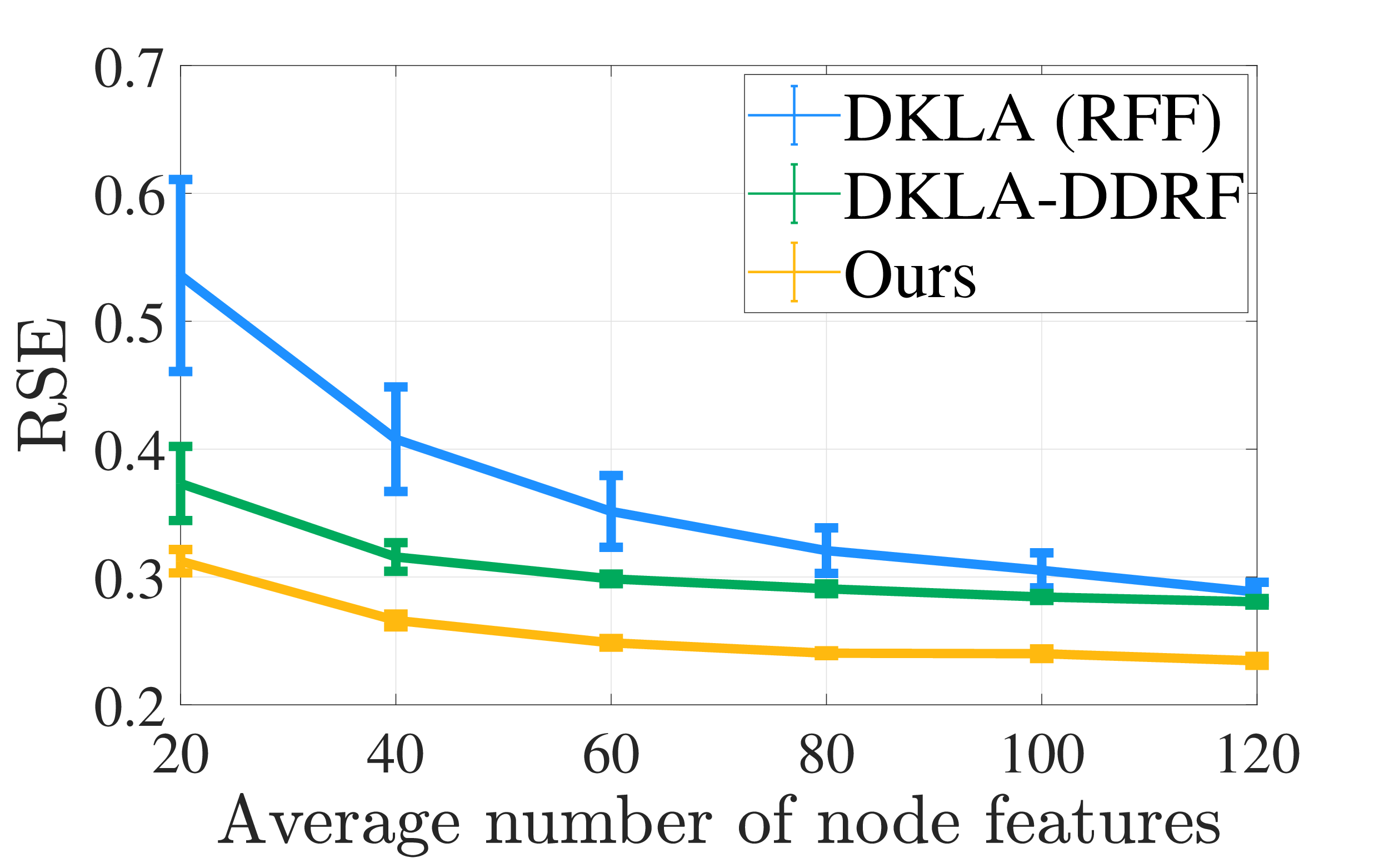}}
	\subfigure[Air Quality]{
		\label{Fig: nonIID.y.2}
		\includegraphics[width=0.31\linewidth]{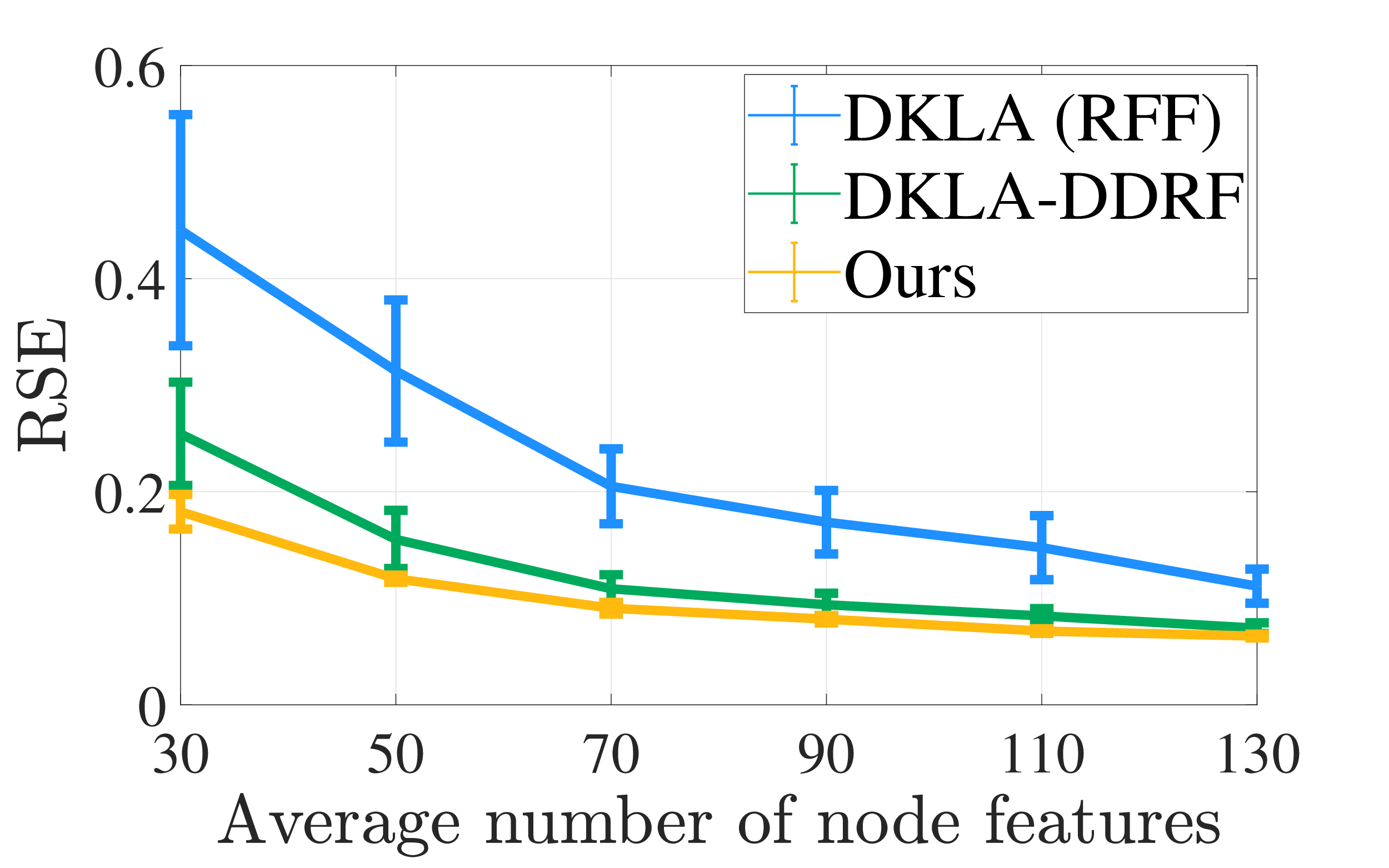}}
	\subfigure[Energy]{
		\label{Fig: nonIID.y.3}
		\includegraphics[width=0.31\linewidth]{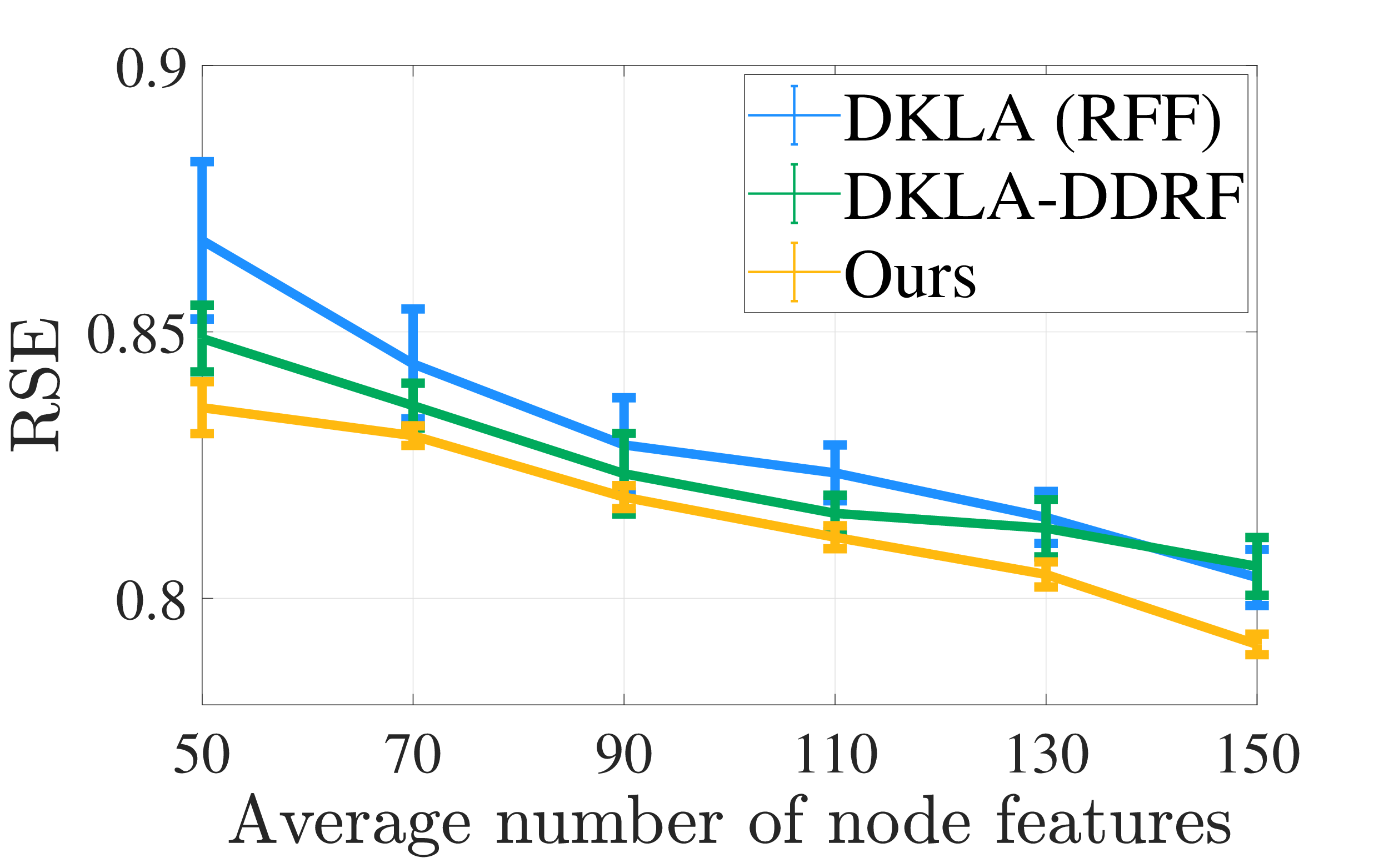}}
	\subfigure[Twitter]{
		\label{Fig: nonIID.y.4}
		\includegraphics[width=0.31\linewidth]{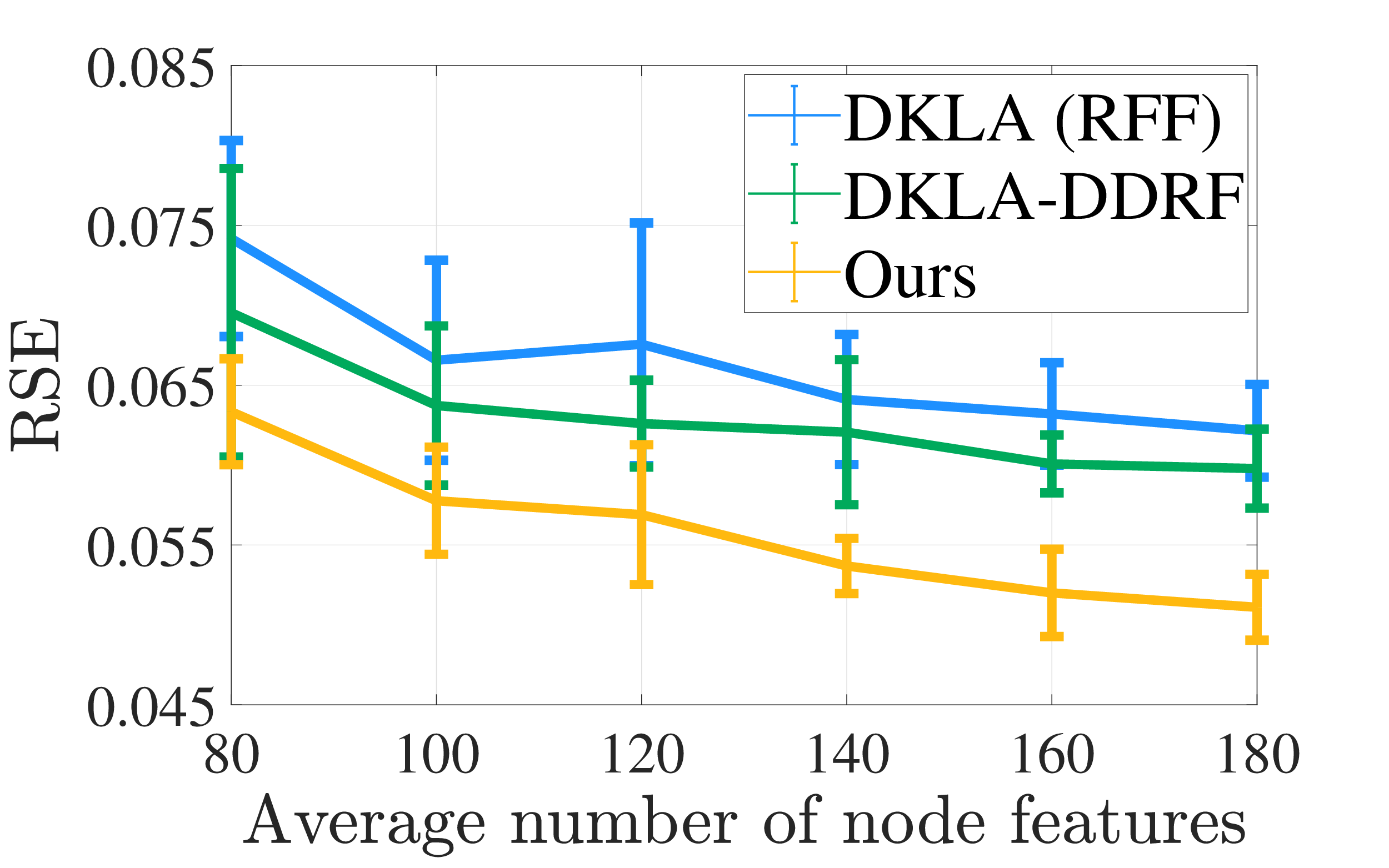}}
	\subfigure[Tom's Hardware]{
		\label{Fig: nonIID.y.5}
		\includegraphics[width=0.31\linewidth]{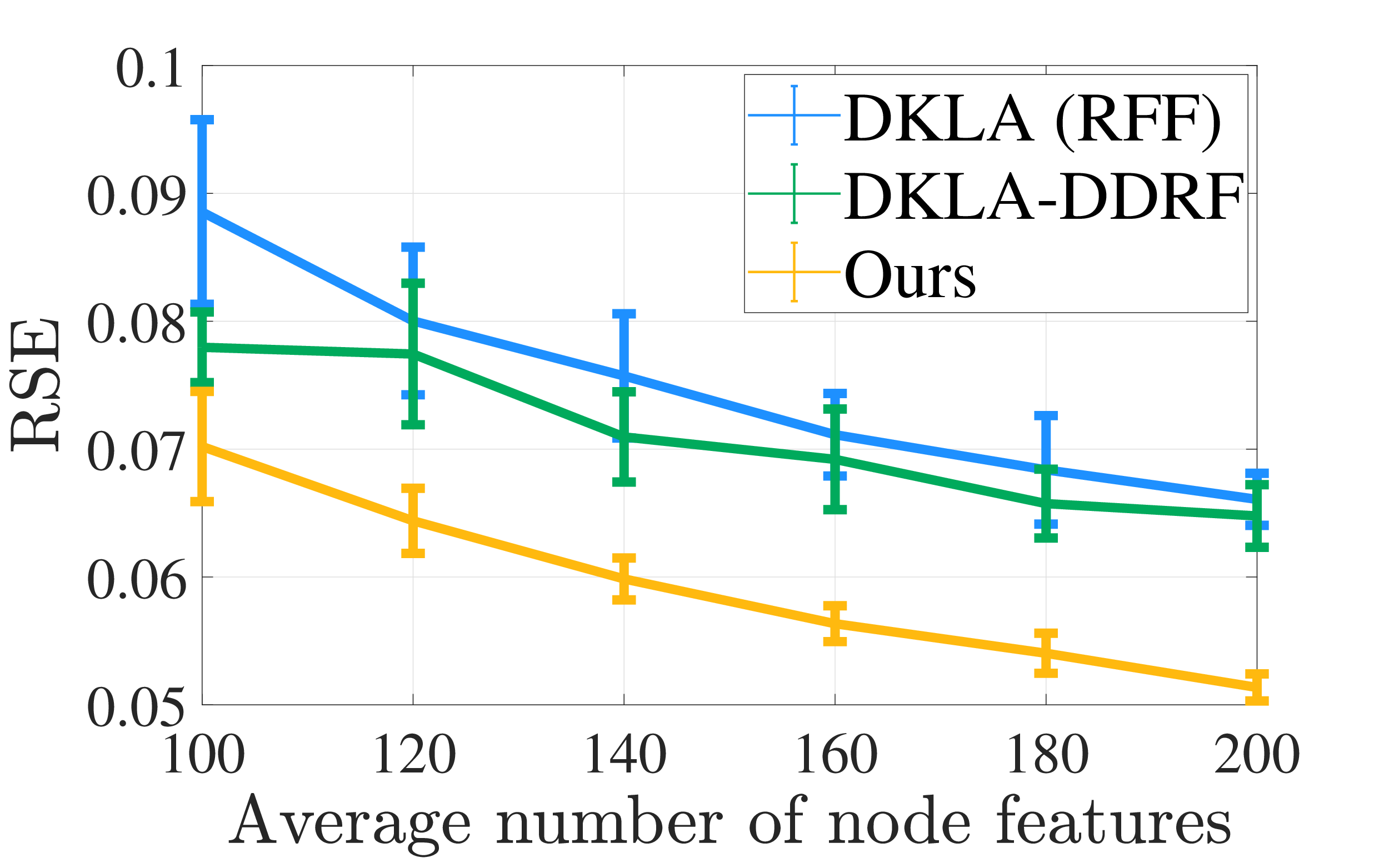}}
	\subfigure[Wave]{
		\label{Fig: nonIID.y.6}
		\includegraphics[width=0.31\linewidth]{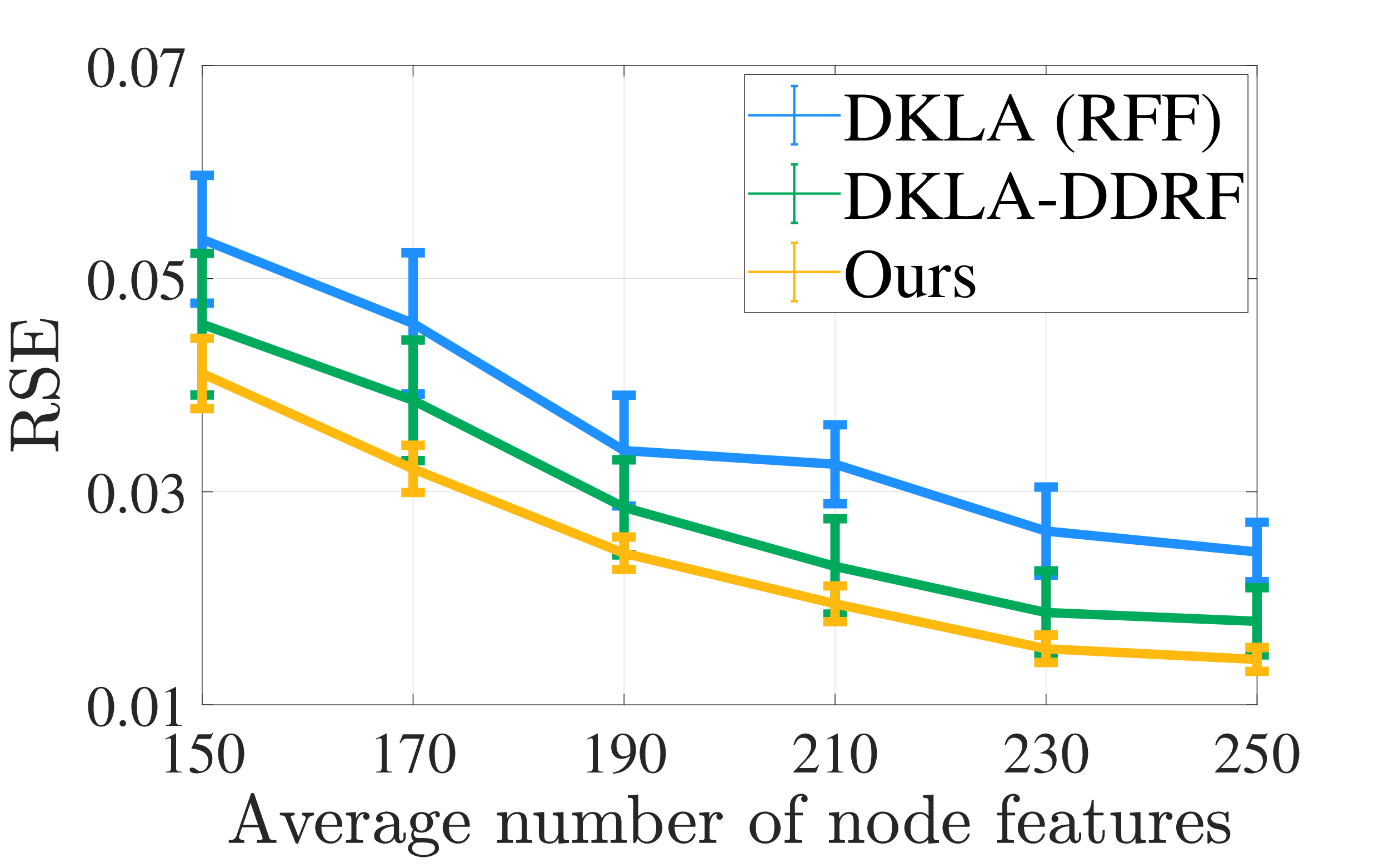}}
	\caption{The RSE (mean$\pm$std) on six test sets versus the average number of features used by each node ($\bar{D}$) in the first non-IID data setting (different $\overline{|y_{j,i}|}$). $J=10$, $|\mathcal{N}_j|=4$, and ten nodes contain the same amount of data $N_j$ and the same number of features $D_j$.}
	\label{Fig: noniid1}
\end{figure*}

\begin{figure*}[htbp]
	\centering  
	\vspace{0cm} 
	\subfigtopskip=2pt 
	\subfigbottomskip=2pt 
	\subfigcapskip=-2pt 
	\subfigure[Houses]{
		\label{Fig: nonIID.x.1}
		\includegraphics[width=0.31\linewidth]{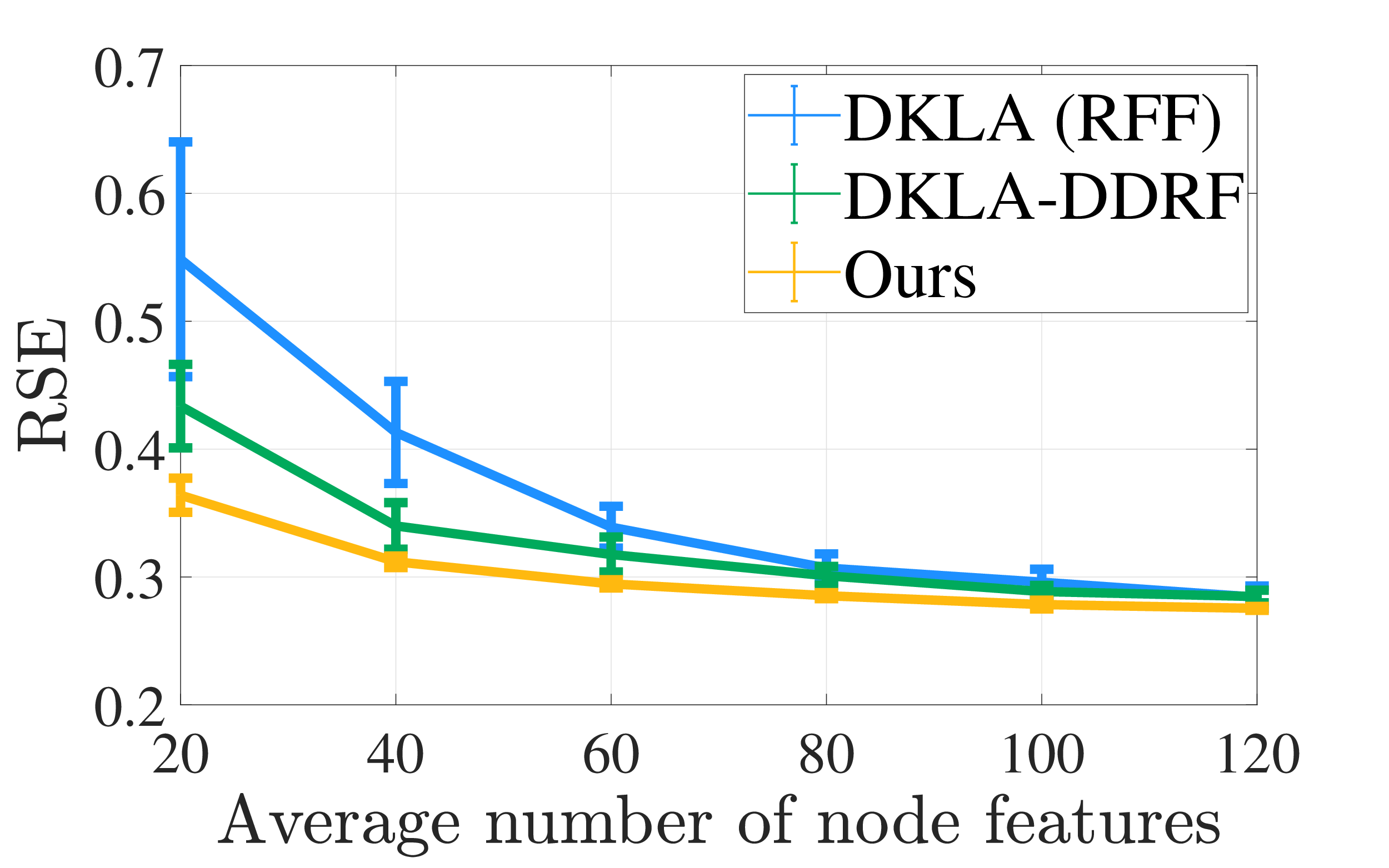}}
	\subfigure[Air Quality]{
		\label{Fig: nonIID.x.2}
 		\includegraphics[width=0.31\linewidth]{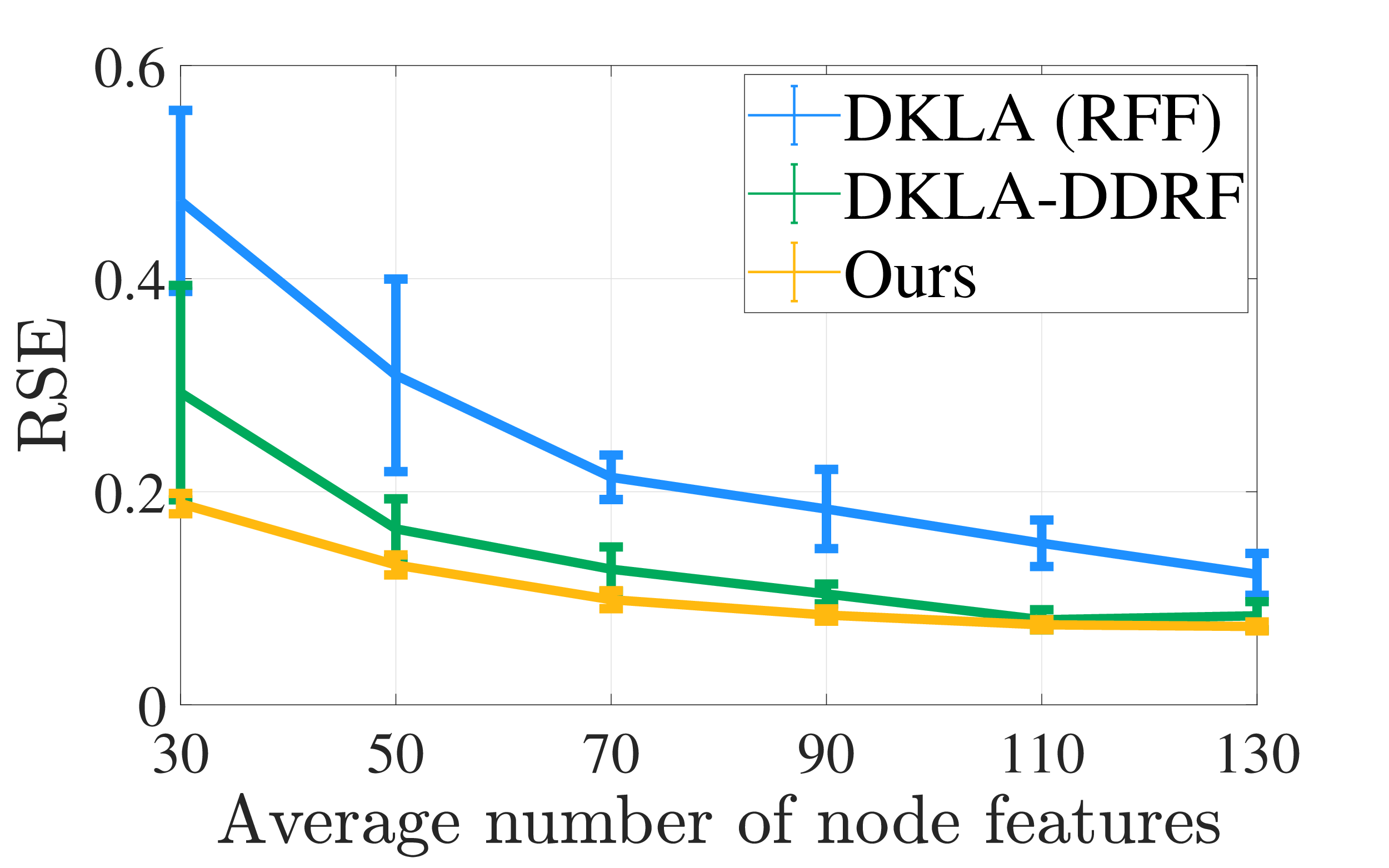}}
	\subfigure[Energy]{
		\label{Fig: nonIID.x.3}
		\includegraphics[width=0.31\linewidth]{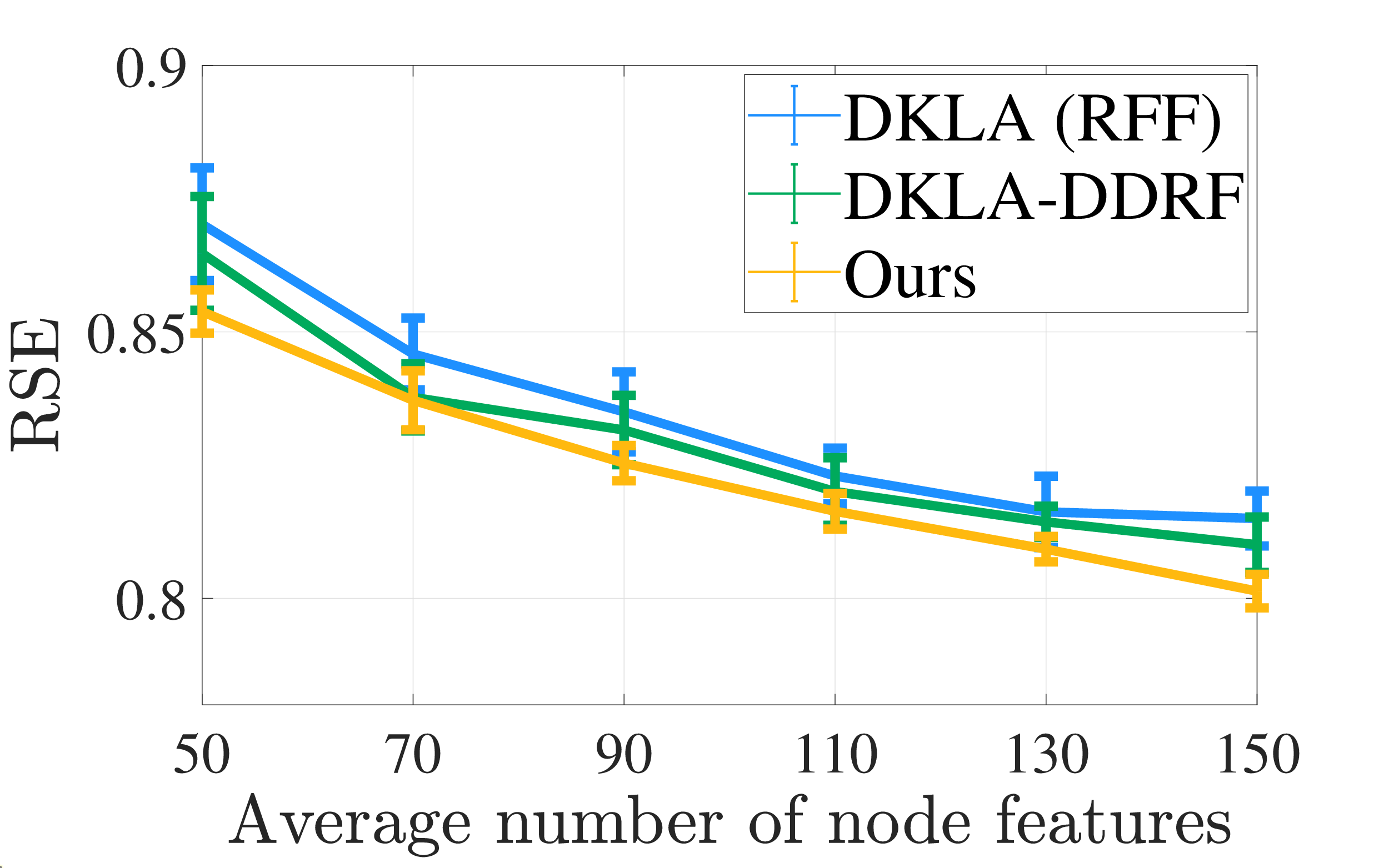}}
	\subfigure[Twitter]{
		\label{Fig: nonIID.x.4}
		\includegraphics[width=0.31\linewidth]{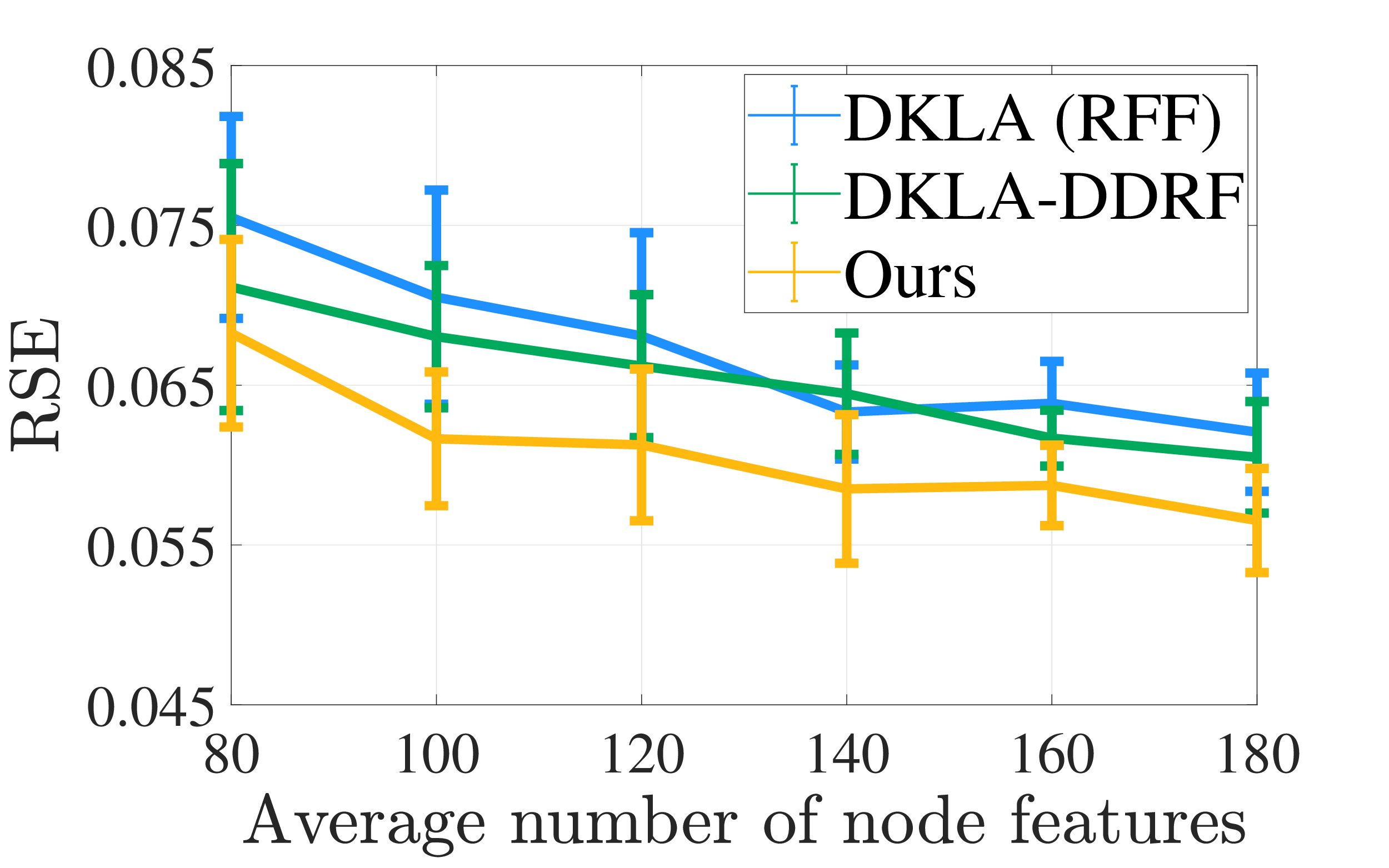}}
    \subfigure[Tom's Hardware]{
    		\label{Fig: nonIID.x.5}
     		\includegraphics[width=0.31\linewidth]{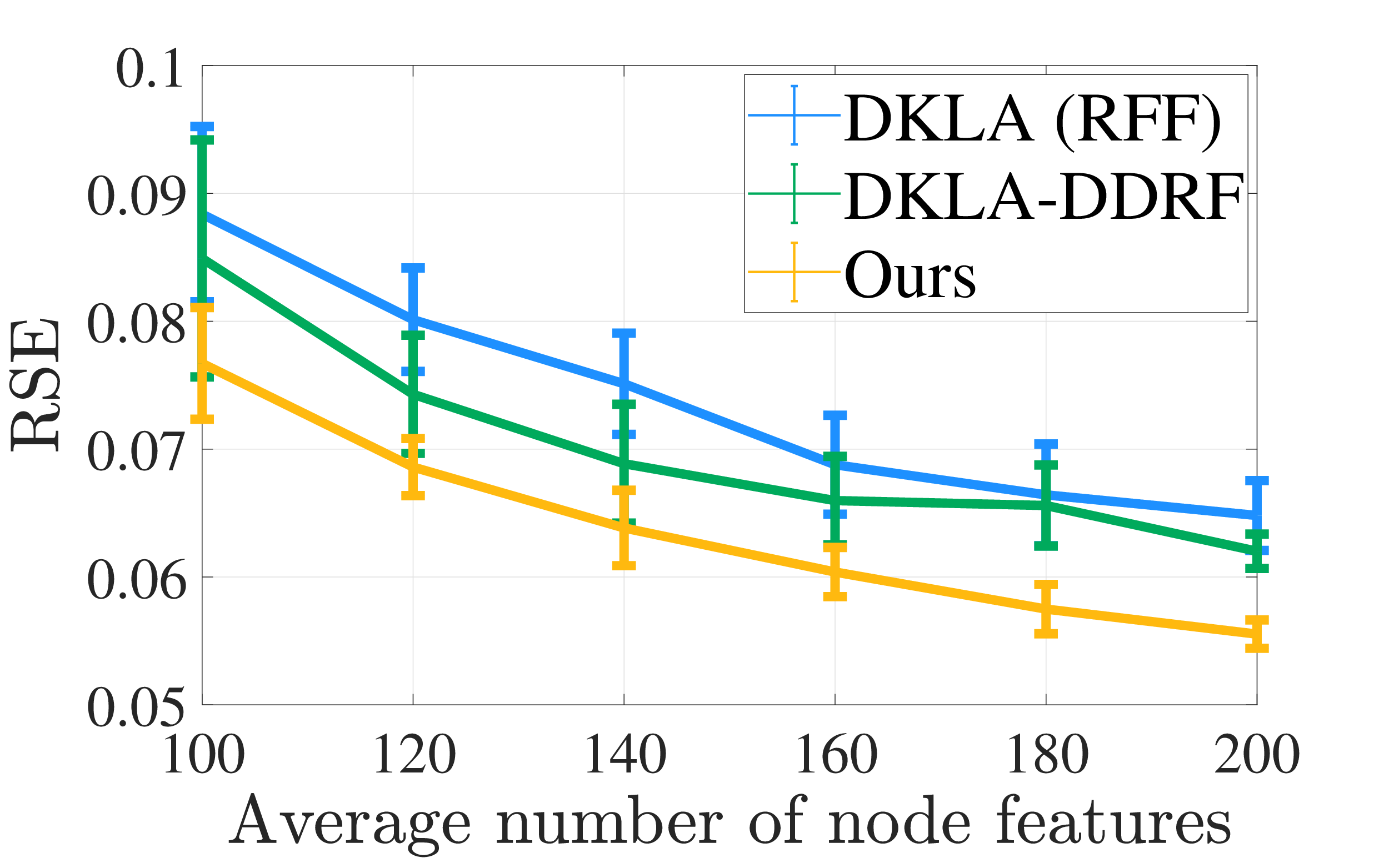}}
	\subfigure[Wave]{
		\label{Fig: nonIID.x.6}
		\includegraphics[width=0.31\linewidth]{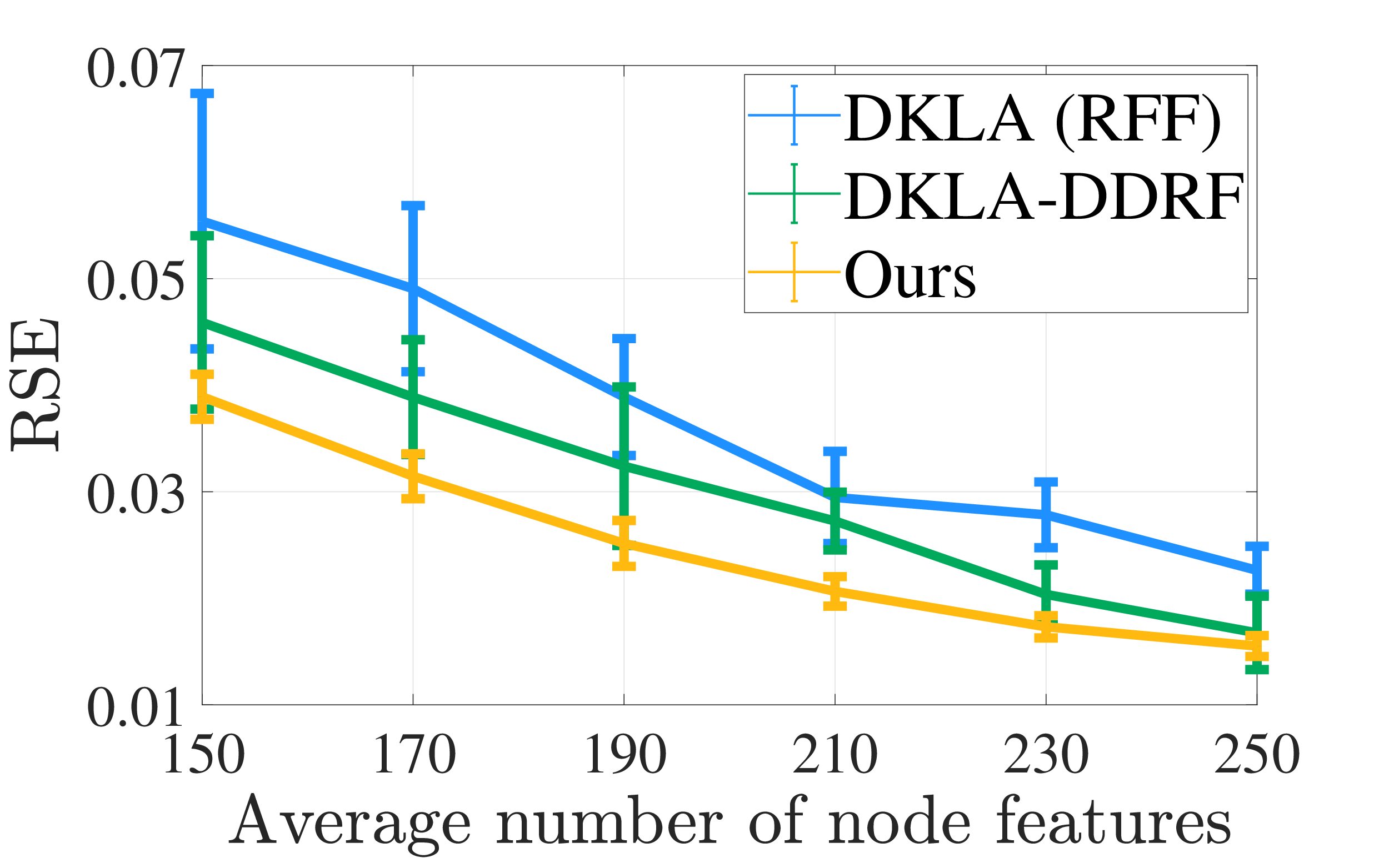}}
	\caption{The RSE (mean$\pm$std) on six test sets versus the average number of features used by each node ($\bar{D}$) in the second non-IID data setting (different $\overline{\|\vx_{j,i}\|_2}$). $J=10$, $|\mathcal{N}_j|=4$, and ten nodes contain the same amount of data $N_j$ and the same number of features $D_j$.}
	\label{Fig: noniid2}
\end{figure*}

\noindent\textbf{Parameter settings.} 
The parameters for centralized KRR, DKLA, and our DeKRR-DDRF are set according to the following approach.

\begin{enumerate}
    \item Centralized KRR: We choose the shift-invariant Gaussian kernel $k(\vx,\vx^{\prime})=\mathrm{exp}\left(\frac{\|\vx-\vx^{\prime}\|_2^2}{2\sigma^2}\right)$ for all experiments. For the regularization parameter $\lambda$ and the bandwidth $\sigma$, we select them by 5-fold cross-validation from candidate sets $\lambda\in\{10^i, i=-8,-7,\ldots,-2\}$, $\sigma\in\{2^i, i=-2,-1,\ldots,2\}$. 
    \item DKLA \cite{xu2021coke}: For the augmented coefficient $\rho$, we set it as $10^{-4}$ and double it after every 200 iterations. 
    \item DeKRR-DDRF (Ours): For the penalty parameters $\{c_{j,\mathrm{nei}}\}_{j=1}^{J}$, we select them by 5-fold cross-validation: $c_{j,\mathrm{nei}}\in\{2^iN, i=-1,0,\ldots,3\},\ \forall j \in\mathcal{V}$, and we set $c_{j,\mathrm{self}}=5c_{j,\mathrm{nei}}=5c_{p,\mathrm{nei}},\ \forall(j,p)\in\mathcal{E}$. For the ratio of $D_0$ to $D_j$, we follow the settings in \cite{shahrampour2018data} and set $D_0/D_j=20$.
\end{enumerate}

\subsection{Experimental Results}
\subsubsection{Non-IID data}

We consider an undirected connected graph with 10 nodes, where each node has 4 neighbors. 
Since each node has the same number of neighbors, that is $|\mathcal{N}_1|=\ldots=|\mathcal{N}_J|$, the total communication costs will be proportional to $D_{\mathrm{all}}\coloneqq J\bar{D} = \sum_{j=1}^J D_j$.
By following the approach outlined in \cite{hong2021communication}, we simulate the non-IID scenario by making the mean value of $\{|y_{j,i}|\}_{i=1}^{N_j}$ or $\{\|\vx_{j,i}\|_2\}_{i=1}^{N_j}$ of the data on each node different. 
We centrally compute $\{|y_{i}|\}_{i=1}^{N}$ or $\{\|\vx_{i}\|_2\}_{i=1}^{N}$, sort them in descending order, and assign $N_j$ data to the $j$-th node in that order.
In these settings, all nodes have the same amount of data and use the same number of random features.

In order to verify that our algorithm can fully exploit the characteristics of each node's data in the non-IID setting, we compare DeKRR-DDRF to a data-independent algorithm that uses the plain RFF (DKLA) and an algorithm that only utilizes data from one node (named DKLA-DDRF).
The mean RSEs are reported in Tab.~\ref{Tab: Difference}, where our method is highlighted in bold, indicating its significant superiority over other methods at the 1\% significance level based on paired t-tests conducted across repeated experiments. Across six real-world datasets, DeKRR-DDRF exhibited average improvements of 25.5\% and 13.6\% in terms of RSE compared to DKLA and DKLA-DDRF, respectively.
The results clearly illustrate that the data-dependent algorithms can achieve lower regression errors than the data-independent one.
Additionally, previous algorithms that require consistent features on all nodes can only use a portion of data for feature selection, making them less effective than our algorithm which can utilize data on all nodes.

\begin{table}[tbp]
\centering
\caption{RSE comparison results of different algorithms in the first non-IID data setting (different $\overline{|y_{j,i}|}$).  The bold values indicate that our method is significantly better than other methods via paired t-test at the 1\% significance level.}
\begin{tabular}{c|c|c|c|c}
\toprule[1.5pt]
Data sets      & $\bar{D}$   & DKLA (RFF) & DKLA-DDRF & Ours  \\
\midrule[1pt]
Houses         & 70  & 0.334      & 0.290       & \textbf{0.244} \\
Air Quality    & 80  & 0.167      & 0.103       & \textbf{0.088} \\
Energy         & 100 & 0.827      & 0.822       & \textbf{0.816} \\
Twitter        & 130 & 0.067      & 0.063       & \textbf{0.054} \\
Tom's Hardware & 150 & 0.076      & 0.069       & \textbf{0.059} \\
Wave           & 200 & 0.033      & 0.028       & \textbf{0.022} \\
\bottomrule[1.5pt]
\end{tabular}
\label{Tab: Difference}
\end{table}

Next, we evaluate DeKRR-DDRF with DKLA and DKLA-DDRF at different communication costs. Fig.~\ref{Fig: noniid1} and Fig.~\ref{Fig: noniid2} show that as the communication costs increase, our DeKRR-DDRF outperforms the other two algorithms in terms of both mean regression error and deviation. Although DKLA-DDRF leverages information from a subset of the data and shows some improvement in the mean regression error compared to DKLA, our algorithm considers the characteristics of all the data comprehensively. Particularly, the DeKRR-DDRF demonstrates superior performance on the \textit{houses}, \textit{twitter}, \textit{Tom's hardware}, and \textit{wave} data sets. Furthermore, because a set of data always has some good features that correspond to it, our data-dependent algorithm can select these features before communication, resulting in more stable results.

\begin{figure}[tbp]
    \centering
    \includegraphics[width=1\linewidth]{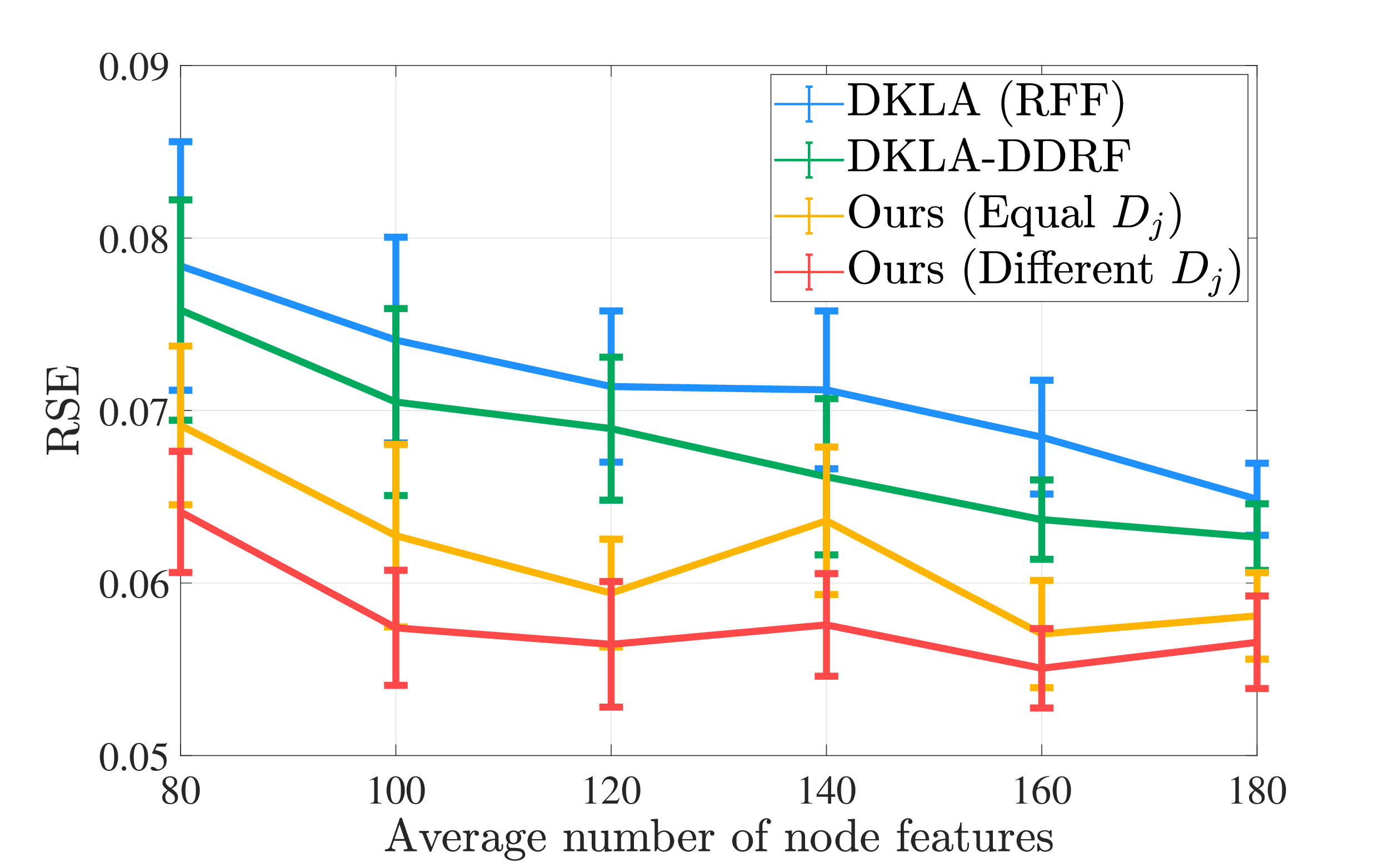}
	\caption{The RSE (mean$\pm$std) on \textit{twitter} data set versus the average number of features used by each node ($\bar{D}$) in the imbalanced data setting, where the $j$th node has $N_j = \frac{2j-1}{100}N$ data $(\sum_{j=1}^J N_j = N)$. $J=10$, $|\mathcal{N}_j|=4$, $\lambda = 10^{-6}$, and $\sigma=4$. Set $D_{j,\mathrm{Ours}}=\bar{D}$ for the equal $D_j$ setting and $D_{j,\mathrm{Ours}}=\sqrt{N}_jJ\bar{D}/\sum_{j=1}^J\sqrt{N}_j$ for the different $D_j$ setting.}
    \label{Fig: imbalanced}
\end{figure}

\subsubsection{Imbalanced Data}
We consider an undirected connected graph with 10 nodes, each with 4 neighbors. And we set the number of data on each node to $N_j = \frac{2j-1}{100}N$, making the $N_j$ of each node significantly different and $\sum_{j=1}^J N_j = N$. Because DKLA requires the same number of features on each node, it is evident that $D_{j,\mathrm{DKLA}}= \bar{D}$. In DeKRR-DDRF, the setting of $D_{j,\mathrm{Ours}}$ is more flexible. An intuitive idea is to assign more features to nodes that have more data, from which we set $D_{j,\mathrm{Ours}}=\sqrt{N}_jJ\bar{D}/\sum_{j=1}^J\sqrt{N}_j$. However, in reality, it can also be flexibly adjusted according to specific needs to prevent the regression ability from being excessively reduced due to a lack of features in small data nodes or the occurrence of redundancy caused by an excessive number of features in big data nodes.
In this way, the total communication costs are consistent with DKLA, but the feature numbers are more appropriate.

\begin{figure*}[tbp]
    \centering
    \includegraphics[width=1\linewidth]{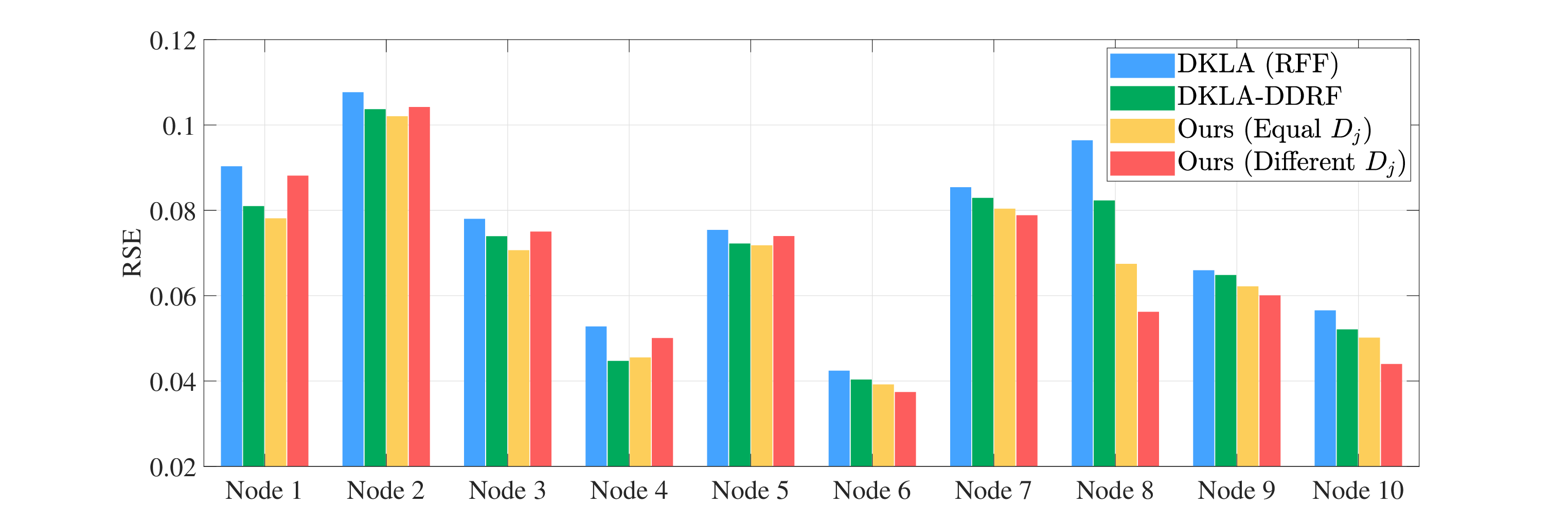}
    \caption{Each node's RSE performance on \textit{twitter} data set in the imbalanced data setting when $\bar{D}=100$. $J=10$, $|\mathcal{N}_j|=4$, $\lambda = 10^{-6}$, $\sigma=4$, and $\sum_{j=1}^J N_j = N$. The number of data from node 1 to node 10 gradually increases, following the setting $N_j=\frac{2j-1}{100} N$. Set $D_{j,\mathrm{Ours}}=\bar{D}$ for the equal $D_j$ setting and $D_{j,\mathrm{Ours}}=\sqrt{N}_jJ\bar{D}/\sum_{j=1}^J\sqrt{N}_j$ for the different $D_j$ setting. By selecting the features and adjusting $D_j$ for each node, the accuracy of big data nodes $(j=6, 7, \ldots, 10)$ is further improved, while the total number of features used by the network remains unchanged.}
    \label{fig: detail}
\end{figure*}

To investigate the benefits, we conduct experiments on the \textit{twitter} with imbalanced data. For our algorithm, we test two settings: equal $D_j$ and different $D_j$, and compare them with DKLA and DKLA-DDRF. In the case of DKLA-DDRF, we utilized the node with the largest amount of data (node 10) for feature selection and broadcast the selected features to all other nodes. As shown in Fig.~\ref{Fig: imbalanced}, when the total number of features used by the four methods is the same, i.e., under the condition of equal communication costs during iterative optimization, our algorithm outperforms DKLA and its DDRF version in terms of regression performance in both settings of $D_j$. And by using different $D_j$, our DeKRR-DDRF achieves adaptive adjustment of the number of features for each node, resulting in further improvement in regression performance. Additionally, user's data is more likely to appear on nodes with more data in reality. Fig.~\ref{fig: detail} illustrates that when the communication costs are unchanged, our algorithm can use the selected features to improve overall regression performance and, in particular, the performance of big data nodes $(j=6, 7, \ldots, 10)$ by assigning these nodes additional features.

\section{Conclusion}
\label{Section: conclusion}
In this paper, a novel decentralized KRR algorithm is proposed. 
Unlike the constraint of aligning parameter vectors in previous works, we instead pursue the consistency of the decision function on each node.
This makes our algorithm more flexible because nodes are no longer required to adhere to the RF consistency premise.
When dealing with more general data types, such as non-IID or imbalanced data, our algorithm can adjust the number of features that each node uses based on the node's computational ability, communication capacity, or data size.
In addition, our algorithm can leverage well-behaved DDRF methods on all nodes to improve the kernel function approximation ability or generalization performance. 
The numerical experimental results demonstrate that, under the condition of using the same number of RFs at each node in the network, our DeKRR-DDRF exhibits lower regression errors compared to other algorithms.
Furthermore, because of the feature selection, our algorithm outperforms other algorithms in terms of stability across multiple experiments.

\section*{Acknowledgments}
The authors would like to thank the anonymous reviewers for their insightful comments.

\bibliographystyle{IEEEtran}
\bibliography{ref}

\end{document}